\pdfoutput=1
\documentclass[11pt]{article}
\usepackage{acl}
\usepackage{times}
\usepackage{latexsym}
\usepackage{amsmath}
\usepackage{amssymb}
\usepackage{graphicx}
\usepackage{booktabs}
\usepackage{multirow}
\usepackage{xcolor}
\usepackage[T1]{fontenc}
\usepackage[utf8]{inputenc}
\usepackage{microtype}
\usepackage{inconsolata}
\usepackage{hyperref}

\newcommand{\pt}{\textsc{pt}}
\newcommand{\lnn}{\textsc{ln}}
\newcommand{\rzero}{\textsc{Zero-shot}}
\newcommand{\rone}{\textsc{Point-only}}
\newcommand{\rtwo}{\textsc{Line-only}}
\newcommand{\rthree}{\textsc{Joint-mixed}}
\newcommand{\rthreeb}{\textsc{Joint-onepass}}
\newcommand{\rfour}{\textsc{Conditioned}}
\newcommand{\rfoursh}{\textsc{Cond.-shuffled}}
\newcommand{\rpipe}{\textsc{Pipeline}}
\newcommand{\rfourn}{\textsc{Cond.-neutral}}
\newcommand{\rlndir}{\textsc{Cond.-fields}}
\newcommand{\rptdir}{\textsc{Cond.-hypo}}
\newcommand{\rseq}{\textsc{Sequential}}

\title{Joint Training Is Not Enough: Conditioned Cross-Granularity Training \\ for Multimodal Document Understanding}

\author{
  Chengguang Gan\textsuperscript{1} \quad
  Yunhao Liang\textsuperscript{2} \quad
  Hanjun Wei\textsuperscript{2} \quad
  Qinghao Zhang\textsuperscript{3} \quad
  Shiwen Ni\textsuperscript{4} \\[0.3em]
  \textsuperscript{1}Independent Researcher, \quad
  \textsuperscript{2}University of Chinese Academy of Sciences \\
  \textsuperscript{3}Department of Information Convergence Engineering,
  Pusan National University, South Korea \\
  \textsuperscript{4}Shenzhen University of Advanced Technology \\[0.25em]
  \textbf{Correspondence:} \href{mailto:chengguangg1024@gmail.com}{chengguangg1024@gmail.com}
}

\begin{document}
\maketitle

\begin{abstract}
The Mutual Reinforcement Effect (MRE) asks whether a fine, span-level and a coarse,
document-level task help each other when one model handles both. We test it in multimodal document
understanding on three corpora, two of receipts and one of scanned business forms, comparing
single-task, joint and \emph{conditioned} training, which puts one granularity's gold output in the
other's prompt during training only. We build Doc-MRE, an annotation layer pairing gold field
extraction (point) with four document-level facets (line), from a three-judge LLM committee under a
pre-registration, validated by blind re-annotation. One predicate, fixed in advance: at a shared
recipe, a regime \emph{reinforces} if it beats the matched single-task model on both granularities.
Mixed joint training, the arrangement prior MRE work assumes, reinforces on no corpus at the main
scale: it is below both single-task models on CORD and trades one granularity for the other on the
two others, as single-task tuning does. Conditioned training reinforces on two of the three, CORD
($+0.5$ point, $+4.8$ line) and the forms corpus ($+7.2$ point, $+11.0$ line), resolvably on the
coarse side and directionally on the fine one, and trades on WildReceipt; at that recipe no
alternative measurably beats it on either side anywhere. Two byte-identical-prompt controls
separate content from format: shuffled conditioning destroys the coarse-side skill but costs the
fine side far less, and a neutral-content control reproduces the whole fine-side gain on
WildReceipt, which is therefore prompt structure but buys nothing resolvable on the other two. On
the forms corpus conditioning buys collapse avoidance: mixed training and the neutral control both
assign the majority semantic label to all $50$ test documents; only conditioning recovers the gold
distribution. Probes find the information decodable under every regime with no resolvable increase
under conditioning.
\end{abstract}

\begin{figure}[t]
\centering
\includegraphics[width=0.98\columnwidth]{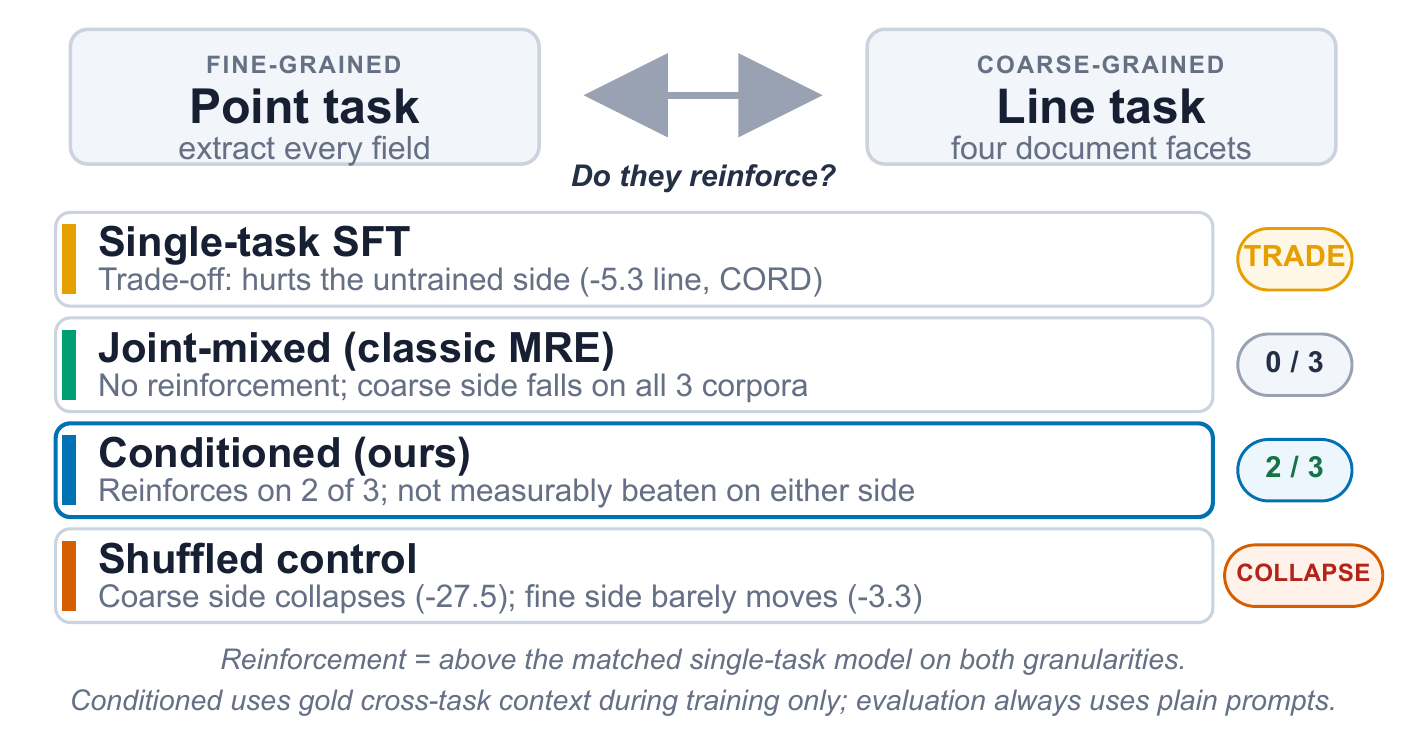}
\caption{Four arrangements of the same task pair, judged by one predicate:
\emph{reinforces} means above the matched single-task model on both granularities. Only
conditioned training does so, on two of three corpora; filling the same prompt with another
document's content collapses the coarse side.}
\label{fig:teaser}
\end{figure}

\section{Introduction}

Document understanding asks one model for answers at two granularities. A receipt, for
example, supports a fine task, extracting every field instance on the page (the \emph{point}
task), and a coarse task, judging document-level properties such as the kind of store and the payment method (the \emph{line} task). The Mutual Reinforcement Effect (MRE) line of work \citep{gan2025m} conjectures that
these granularities reinforce each other when a single model handles both: knowing the
document type constrains which fields to expect, and extracted fields settle document-level
questions that would otherwise require guessing.

Whether this reinforcement actually arrives, and what training arrangement produces it, has not
been tested with matched controls in the document setting. We do so here on three corpora, two of
receipts and one of scanned business forms \citep{jaume2019funsd}. We fix the predicate once, in
the form the conjecture asserts it: a regime \emph{reinforces} if it is above the \emph{matched
single-task model} on both granularities, and \emph{trades} if it is above on one and below on
the other. Every claim below is stated against that reference and against the untuned model, we never switch reference within a comparison, and every regime is trained with one shared recipe whose sensitivity we then measure.

Mixed joint training, the arrangement prior MRE work assumes, reinforces on no corpus at the main scale (Figure~\ref{fig:teaser}). On CORD it is below both single-task models ($-2.6$ point, $-0.5$
line); on WildReceipt and FUNSD it trades, buying the fine side and losing the coarse side
resolvably ($-2.5$ and $-4.0$). Single-task tuning trades harder still, costing the untrained
granularity $5.3$ line points on CORD with every prediction still well formed. Conditioned
training, in which the prompt for each task carries the gold output of the other during training
only while both prompts are plain at test, reinforces on two of the three corpora, CORD ($+0.5$
point, $+4.8$ line) and FUNSD ($+7.2$ point, $+1.9$ once both arms must parse, and $+11.0$ line), with the coarse side resolvable in both and the fine side directional; on WildReceipt it trades, its coarse side sitting $1.2$
points below line-only tuning within the paired interval. At that recipe no alternative beats it measurably on either side of any corpus.

Four controls pin down what drives the gain, and they disagree across corpora. Shuffled conditioning, identical in format but filled from a different document, destroys the coarse-side skill and costs the fine side far less, $3.3$ points on CORD against $27.5$. Neutral conditioning, identical in format but with uninformative placeholders, is the decisive one: on WildReceipt it reproduces the
whole fine-side gain, so that gain is prompt structure rather than content, whereas on CORD and
FUNSD it buys nothing resolvable and the real content is worth a further $+4.3$ and $+17.0$
line points. On FUNSD mixed training and the neutral control both collapse the semantic facet to its majority class for all $50$ test documents at that recipe, in three seeds, while conditioning recovers the gold distribution: the sharpest content effect we observe. A step-matched control doubling
single-task epochs moves nothing, and three retraining seeds hold the gains fixed.

We also ask what conditioned training changes inside the model, and report it as descriptive. Layer-wise probes with selectivity controls \citep{hewitt2019designing} find the cross-granularity information decodable under every regime, zero-shot included, with no increase under conditioning that we can resolve, so the gain is not injection.
Input-side interventions and gradient attribution are consistent with a readout-level account
but do not establish one, and three of the four instruments share a prompt-format confound with the effect they measure, which the neutral control bounds for one of them.

Our contributions are:
\begin{itemize}\setlength\itemsep{0.15em}
\item \textbf{A controlled negative and a partial repair.} Under one predicate fixed in advance, mixed joint training reinforces on none of three corpora at the main scale and single-task tuning trades one granularity for the other, while conditioned training reinforces on two, trades on the third, and at that recipe is never measurably beaten on either side.
\item \textbf{An analysis battery reported with its nulls.} Four instruments
(selectivity-controlled probes, input interventions, gradient attribution against a lexical
null, and a modality ablation), each with its control. Conditioning adds no decodable
cross-granularity information over zero-shot while mixed co-training removes some; the input-side instruments are consistent with a readout-level account without establishing it, and three of the four run in formats only the conditioned model was trained on.
\item \textbf{Doc-MRE.} An annotation layer over 991 CORD receipts, 400 WildReceipt receipts
and 199 FUNSD business documents, pairing gold field extraction with four document-level facets
and directional evidence links, annotated by a three-judge LLM committee under a pre-registration and validated by blind human re-annotation.
\end{itemize}

\section{Related Work}

\paragraph{Mutual Reinforcement Effect.} The MRE line conjectures bidirectional gains from
jointly modeling fine and coarse IE tasks, and its multimodal instantiation M-MRE
\citep{gan2025m} extends the setting to vision-language models, reporting that transfer
between granularities is not automatic. We move the question to document understanding, where
the two granularities share one page and gold evidence links between them can be annotated,
and we replace the joint-training assumption with a controlled comparison of training
regimes. Our results revise the default reading of this line: on CORD mixed training produces no
reinforcement, and where reinforcement appears at all it is larger and more consistent when
trained for explicitly through conditioning.

\begin{figure*}[t]
\centering
\includegraphics[width=0.92\textwidth]{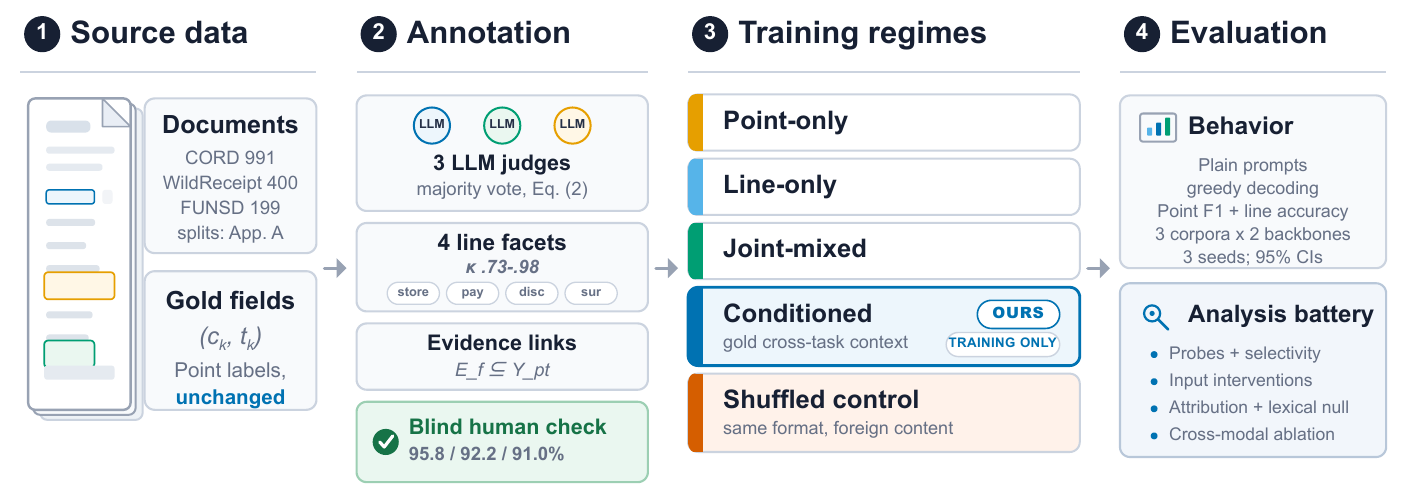}
\caption{Overview. (1) Source documents keep their gold field annotations as point labels.
(2) A three-judge LLM committee adds four document-level facets and directional evidence
links by majority vote; blind human re-annotation validates the labels at 95.8\%, 92.2\% and 91.0\% agreement on the three corpora. (3) Five training regimes arrange the same task pair differently; only
\rfour{} sees gold cross-granularity context, and only during training; the shuffled control
keeps its format but swaps in another document's content. (4) All models are evaluated with plain prompts on three corpora, two backbones and three seeds, then analyzed with a four-instrument battery.}
\label{fig:overview}
\end{figure*}

\paragraph{Learning with privileged information.} Conditioned training instantiates learning
using privileged information \citep{vapnik2009new,lopez2015unifying} in prompt space:
training-time inputs carry signal unavailable at test time. Context distillation \citep{snell2022learning} internalizes prompt content into weights, intermediate-task transfer \citep{phang2018sentence} orders tasks rather than conditioning them, and rationale-based distillation \citep{hsieh2023distilling} conditions on teacher-written explanations. Our setting differs in structure: the privileged signal is the gold output of a second task
the same model must also learn, the conditioning runs in both directions at once, and the
teacher string is a label the model itself produces at test time, not an external rationale.
Our controls speak to this. A wrong context is far worse than none on the coarse side, and a byte-identical neutral arm shows the transferred quantity to be the content itself on CORD but the prompt structure on WildReceipt. A distillation reading predicts neither dissociation.

\paragraph{Negative transfer in multi-task learning.} That naive task mixing can hurt is
documented for both classic multi-task learning \citep{caruana1997multitask,standley2020tasks}
and gradient-surgery remedies \citep{yu2020gradient}. Our results add a diagnosis for the document-understanding instance (the information
survives; its use does not) and a remedy that changes only the training data.

\paragraph{Document understanding and probing.} CORD \citep{park2019cord}, WildReceipt \citep{sun2021spatial} and FUNSD \citep{jaume2019funsd} supply gold field annotations; document-AI systems such as
LayoutLMv3 \citep{huang2022layoutlmv3} and Donut \citep{kim2022ocr} compete on
extraction itself; we hold one backbone fixed, compare training arrangements, and add
facet labels and evidence links on top of the existing benchmarks. Our analysis
tools follow the probing literature, linear probes \citep{alain2016understanding} with control tasks
\citep{hewitt2019designing}, the caution that decodable is not used \citep{elazar2021amnesic},
gradient saliency \citep{simonyan2013deep}, and LLM-committee annotation \citep{zheng2023judging}.

\section{The Doc-MRE Benchmark}
\label{sec:benchmark}

\subsection{Task pair}

Each instance is a document image $I$ with gold fields
$Y_{\pt}=\{(c_k,t_k)\}_{k=1}^{m}$, category-text pairs from the source dataset, and four
document-level facet labels $y_{\lnn}=(y_f)_{f\in\mathcal{F}}$ with
$\mathcal{F}=\{\texttt{store\_type},\allowbreak\texttt{payment\_method},\allowbreak
\texttt{has\_discount},\allowbreak\texttt{has\_surcharge}\}$.
\texttt{store\_type} is semantic (five classes, judged holistically);
the other three are structural, anchored in specific field types. The anchoring does not
make the line task a lookup over the point output: a symbolic oracle mapping gold fields
to the structural facets reaches line accuracy $.811$ on CORD and $.690$ on WildReceipt
(Appendix~\ref{app:tables}), below zero-shot and below every 8B regime trained on the line task except the shuffled control.

The \emph{point} task maps $(I,x_{\pt})\!\to\!\hat{Y}_{\pt}$, a JSON list of field instances,
scored by exact-match multiset F1 after whitespace and case normalization:
\begin{align}
\label{eq:f1}
\text{F1} &= \frac{2PR}{P+R},\\
P = \frac{|\hat{Y}_{\pt}\sqcap Y_{\pt}|}{|\hat{Y}_{\pt}|}&,\qquad
R = \frac{|\hat{Y}_{\pt}\sqcap Y_{\pt}|}{|Y_{\pt}|},\nonumber
\end{align}
where $\sqcap$ is multiset intersection over $(c,t)$ pairs. The \emph{line} task maps
$(I,x_{\lnn})\!\to\!\hat{y}_{\lnn}$, scored by macro accuracy over facets,
$\mathrm{Acc}=\tfrac{1}{|\mathcal{F}|}\sum_{f}\mathbf{1}[\hat{y}_f=y_f]$.

\subsection{Construction}

Point labels are the source datasets' gold annotations, unchanged. Line labels and
\emph{directional evidence links}, the set $E_f\subseteq Y_{\pt}$ of field instances that
evidence facet $f$, are produced by a three-judge LLM committee (GPT-5.5, Claude Sonnet,
Gemini Pro) under a fixed schema, aggregated by majority vote. Writing $y_f^{(j)}$ for judge $j$'s label,
\begin{equation}
\label{eq:vote}
y_f = \arg\max_{v}\ \textstyle\sum_{j=1}^{3}\mathbf{1}\!\left[y_f^{(j)}=v\right],
\end{equation}
kept only when the winning count is at least two, and the evidence link set keeps the field
instances named by at least two judges,
$E_f=\{e\in Y_{\pt} : |\{j: e\in E_f^{(j)}\}|\ge 2\}$. Samples where any facet lacks a majority are dropped (nine in CORD, none in WildReceipt).
Committee agreement is high: per-facet majority 99--100\% and Fleiss $\kappa$
$.846$--$.978$ on CORD, $.830$--$.948$ on WildReceipt. Evidence-link pairwise Jaccard ranges $.49$--$.98$, so links are used only for analysis, never as training targets.

\paragraph{Blind human validation.} Three annotators independently re-labeled 100 unseen receipts from the CORD training pool without access to committee labels, following a written guide (Appendix~\ref{app:guide}); the human label per facet is their majority vote. Agreement with the committee majority is 95.8\% over 400 CORD facet
decisions ($\kappa$ .836--.980); the 17 disagreements concentrate in
\texttt{store\_type} boundary cases, consistent with that facet's lower committee $\kappa$.
The same protocol reaches 92.2\% ($\kappa$ .69--.95) on the full WildReceipt test set and 91.0\% on FUNSD's, the residual in \texttt{has\_surcharge}, where photographed tax and service lines are harder to read. The nine no-majority CORD samples stay out of every split and ship separately with human-consensus labels. The CORD sample comes from the training pool, so it bounds annotation noise rather than test-set noise; on the other two the validated set is the test set. The resulting label noise, roughly 4\%, is identical for all regimes, since every comparison is paired on the same labels.

\section{Training Regimes}
\label{sec:regimes}

All regimes fine-tune the same backbone (Qwen3-VL-8B-Instruct \citep{bai2025qwen3}; 4B in
Appendix~\ref{app:4b}) with identical LoRA \citep{hu2022lora} and optimization settings
(Appendix~\ref{app:train}), on the same 794 CORD training receipts, with loss on answer
tokens only. Writing $\theta$ for the adapted parameters, the regimes differ only in how the
two tasks are arranged:
\begin{align}
\mathcal{L}_{\textsc{pt}} &= -\log p_\theta\!\left(Y_{\pt}\mid I, x_{\pt}\right)
\tag{\rone}\\
\mathcal{L}_{\textsc{ln}} &= -\log p_\theta\!\left(y_{\lnn}\mid I, x_{\lnn}\right)
\tag{\rtwo}\\
\mathcal{L}_{\textsc{mix}} &= \mathcal{L}_{\textsc{pt}} + \mathcal{L}_{\textsc{ln}}
\tag{\rthree}\\
\mathcal{L}_{\textsc{one}} &= -\log p_\theta\!\left(y_{\lnn}, Y_{\pt}\mid I, x_{\textsc{joint}}\right)
\tag{\rthreeb}
\end{align}
\rthree{} is the classic MRE arrangement, both tasks as separate examples in one mixed
dataset; \rthreeb{} asks for both outputs in a single pass. The conditioned regime places the
gold output of the \emph{other} granularity in the prompt, during training only:
\begin{align}
\label{eq:r4}
\mathcal{L}_{\textsc{cond}} =
&-\log p_\theta\!\left(Y_{\pt}\mid I, x_{\pt}, g(y_{\lnn})\right) \nonumber\\
&-\log p_\theta\!\left(y_{\lnn}\mid I, x_{\lnn}, h(Y_{\pt})\right),
\end{align}
where $g$ renders the facet labels as a stated hypothesis and $h$ renders the field list as
context, both with an instruction to trust the image over the context
(Appendix~\ref{app:prompts}). At evaluation time every model, including \rfour{}, receives
the same plain prompts; no gold context is ever available at test time.

Two controls keep the conditioning format and change its content. For sample $i$, \rfoursh{} trains on $g(y_{\lnn}^{\pi(i)})$ and $h(Y_{\pt}^{\pi(i)})$ from a different document $\pi(i)$, so the slot is filled but wrong; \rfourn{} fills the same slots with uninformative placeholders, so the template is byte-identical and the slot asserts nothing. The second is what bounds a format-level contribution, since the first changes both the content and its truth.
\rone{} and \rtwo{} see half as many examples as \rthree{} and \rfour{} (equal per-task
supervision); a step-matched control doubling their epochs changes point F1 by $-0.1$ and line
accuracy by $-0.5$ points, so this budget difference explains none of what follows.

\section{Main Results}
\label{sec:results}

\begin{figure*}[t]
\centering
\includegraphics[width=0.94\textwidth]{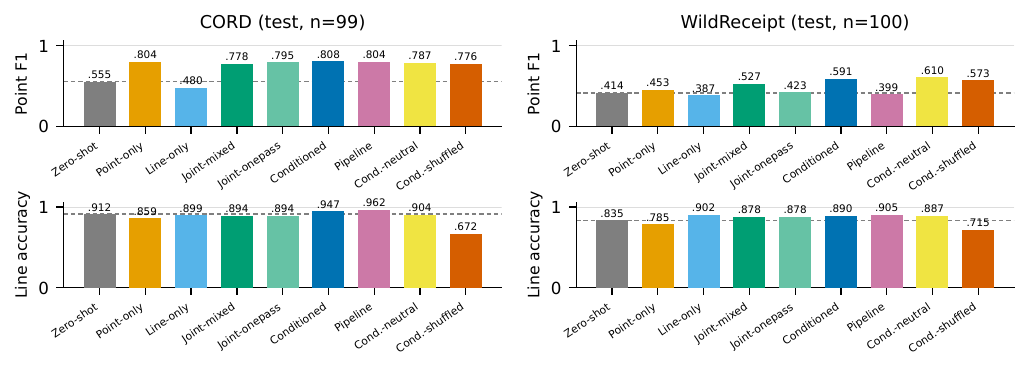}
\caption{Point F1 (top) and line accuracy (bottom) on CORD and WildReceipt test sets, 8B
backbone, plain prompts at evaluation. Dashed line marks zero-shot. No single-forward-pass baseline
measurably beats \rfour{} on either side. The two rightmost bars are controls: \rfourn{}
fills the same slots with uninformative placeholders, \rfoursh{} with another document's
content.}
\label{fig:main}
\end{figure*}

\begin{table}[t]
\centering\footnotesize\setlength{\tabcolsep}{2pt}
\begin{tabular}{lcc|cccc}
\toprule
& & & \multicolumn{4}{c}{Line acc by facet} \\
Regime & Point & Line & store & pay & disc & sur \\
\midrule
majority & -- & .639 & .424 & .636 & .899 & .596 \\
\rzero{} & .555 & .912 & .869 & .919 & .909 & .949 \\
\rone{} & \textbf{.804} & .859 & .828 & .808 & .879 & .919 \\
\rtwo{} & .480 & .899 & .869 & .838 & .919 & .970 \\
\rthree{} & .778 & .894 & .838 & .808 & .939 & .990 \\
\rthreeb{} & \textbf{.795} & .894 & .848 & .798 & .939 & .990 \\
\rfour{} & \textbf{.808} & \textbf{.947} & .899 & .949 & .960 & .980 \\
\rpipe{} & \textbf{.804} & \textbf{.962} & .929 & .949 & .980 & .990 \\
\rlndir{} & .793 & \textbf{.944} & .879 & .960 & .949 & .990 \\
\rptdir{} & .774 & .891 & .848 & .808 & .919 & .990 \\
\rseq{} pt$\to$ln & \textbf{.801} & .896 & .869 & .808 & .929 & .980 \\
\rseq{} ln$\to$pt & \textbf{.804} & .894 & .869 & .828 & .919 & .960 \\
\midrule
\rfourn{} & .787 & .904 & .889 & .798 & .949 & .980 \\
\rfoursh{} & .776 & .672 & .586 & .636 & .899 & .566 \\
\bottomrule
\end{tabular}
\caption{CORD test (n=99, seed 0) with per-facet line accuracy. Bold marks, in the two headline columns only,
the best value in the column and every value whose paired difference from it has a 95\% CI
including zero (intervals in Appendix~\ref{app:tables}); nothing else is bolded and per-facet
columns are unbolded. Every per-facet value is $k/99$, so one receipt is $1.01$ points.
\rpipe{} (Appendix~\ref{app:pipe}) conditions each task on the other's stage-1 predicted output
at train and test, so unlike every other row it needs two forward passes.}
\label{tab:main}
\end{table}

\paragraph{Single-task tuning damages the other granularity.} Relative to zero-shot,
\rone{} drops line accuracy from $.912$ to $.859$ ($-5.3$ points, CI $[-8.8,-2.0]$) and
\rtwo{} drops point F1 from $.555$ to $.480$ ($-7.6$, $[-11.0,-4.4]$). Single-task tuning also specializes
the output format (\rone{} collapses to $.032$ F1 in an unseen joint output format), so we
checked whether the damage is that artifact. On the coarse side it is not: all $99$ of
\rone{}'s line outputs parse with legal values, so the loss lies entirely in the answers. On
the fine side it partly is: over the receipts both models parsed, \rtwo{}'s loss is $-5.5$
$[-8.3,-2.9]$ rather than $-7.6$, and on WildReceipt it does not survive the correction
($-0.6$, $[-1.9,+0.8]$), so we scope fine-side damage to CORD. Both survive retraining at $2{\times}10^{-5}$ (Appendix~\ref{app:train}), and Section~\ref{sec:analysis} shows the information survives where the behaviour does not.

\paragraph{Joint training recovers neither side.} \rthree{}, the arrangement prior MRE work
assumes, sits below \rone{} on point ($.778$ vs $.804$) and below \rtwo{} on line ($.894$ vs
$.899$, Table~\ref{tab:main}). The single-pass variant \rthreeb{} behaves the same, and retraining all three arms at $2{\times}10^{-5}$ reproduces it ($-3.3$ point, $-0.3$ line). Co-training prevents the worst of the cross-task damage but produces no reinforcement here; the pair already interferes before any training (Appendix~\ref{app:pilot}).

\paragraph{Conditioned training reinforces, asymmetrically.} \rfour{} gains $+4.8$ points on
line over \rtwo{} ($[2.3,7.3]$), $+5.3$ over \rthree{} ($[2.5,8.1]$), and $+3.5$ over
zero-shot ($[1.0,6.1]$), while matching \rone{} on point ($+0.5$, $[-1.8,+2.6]$) and beating
\rthree{} there ($+3.0$, $[1.1,5.2]$). Three full retraining seeds reproduce the pattern (Table~\ref{tab:robust}).

\paragraph{Where the line gain comes from, and what it is not.} Among the five pre-registered regimes \rfour{} is best on three of four facets and within one receipt on \texttt{has\_surcharge} (Table~\ref{tab:main}); \rlndir{}, post hoc, is one receipt above it on \texttt{payment\_method}. The gain is uneven: \texttt{payment\_method} alone carries $11.1$ of the $19.2$ facet points separating \rfour{} from \rtwo{}, where \rtwo{} is itself $8.1$ below zero-shot, so more than half of the macro figure is a baseline's self-damage on one facet.

This matters because our symbolic oracle recovers three of the four facets from the gold field
list ($.929$, $.919$, $.970$) and \texttt{store\_type} only at its majority class. An account in which conditioning teaches that rule rather than a cross-granularity dependency predicts the pattern above, so we test \texttt{store\_type} alone (Appendix~\ref{app:tables}). On CORD it is not excluded, \rfour{}'s advantage over zero-shot, \rtwo{} and the neutral control being $+3.0$, $+3.0$ and $+1.0$, none resolvable; on WildReceipt it is, $+5.0$ over both \rthree{} and \rfourn{} ($[1.0,10.0]$). The positive claim therefore lives on the semantic facet rather than in CORD's macro average, as the third domain makes explicit.

\paragraph{One direction carries the gain, and it is the one pointing at the headroom.}
Conditioning runs in both directions at once, so we split it: \rlndir{} conditions only the
line task on the gold field list, \rptdir{} only the point task on the gold facets, each
leaving the other prompt plain. On each dataset exactly one half does the work, and it is the
half feeding the side that gains. On CORD it is fields-to-facets: \rlndir{} reaches line
$.944$, $+5.1$ over \rthree{} ($[2.3,8.1]$) and indistinguishable from full conditioning
($-0.3$, $[-1.8,+1.5]$), while \rptdir{} adds nothing to point ($-0.4$, $[-2.5,+1.8]$). On
WildReceipt it is the reverse: \rptdir{} reaches point $.596$, $+7.0$ over \rthree{}
($[3.2,10.9]$) and indistinguishable from full conditioning ($-0.5$, $[-5.0,+3.8]$), while
\rlndir{} adds nothing to line ($-2.0$, $[-5.2,+1.0]$). Half the recipe is therefore
sufficient on either dataset, and the useful half is not the one the zero-shot pilot
identified (Appendix~\ref{app:pilot}).

\paragraph{Two controls separate content from format.} \rfoursh{} keeps the prompt shape and fills it from a
different receipt. Its line accuracy collapses to $.672$, $27.5$ points below \rfour{}
(Table~\ref{tab:main}), and to $.715$ on WildReceipt, so on the coarse side a wrong context is far worse than none. The fine side is far less affected: $.776$ on CORD, $3.3$ points below \rfour{} ($[-5.1,-1.5]$), and $.573$ on WildReceipt, $1.8$ below it within the interval ($[-7.1,+3.6]$). \rfourn{}, whose template is byte-identical, is what separates content from format.

The two datasets dissociate. On CORD the neutral slot buys nothing resolvable while \rfour{}
exceeds it by $+4.3$ line ($[1.5,6.8]$) and $+2.2$ point ($[0.6,4.0]$), so content carries the
gain. On WildReceipt the fine side reverses: \rfourn{} reaches point $.610$, $+8.3$ over
\rthree{} ($[4.2,12.8]$) against \rfour{}'s $+6.4$, and \rfour{} is indistinguishable from it ($-1.9$, $[-7.0,+2.8]$), as is \rptdir{}. There prompt structure does the work.

\begin{table}[t]
\centering\footnotesize\setlength{\tabcolsep}{1.5pt}
\begin{tabular}{lcc}
\toprule
& Point F1 & Line Acc \\
\midrule
\multicolumn{3}{l}{\emph{Seeds 0/1/2 (CORD test, 8B)}} \\
\rfour{} & \textbf{.808}/.799/.815 & \textbf{.947}/.947/.944 \\
\rthree{} & .778/.795/.782 & .894/.889/.884 \\
\rone{} & .804/.784/.788 & .859/.866/.866 \\
\midrule
\multicolumn{3}{l}{\emph{Dev split (n=98, 8B)}} \\
\rzero{} / \rone{} & .559 / .822 & .913 / .842 \\
\rtwo{} / \rthree{} & .473 / .787 & .883 / .870 \\
\rfour{} & \textbf{.824} & \textbf{.941} \\
\midrule
\multicolumn{3}{l}{\emph{4B backbone (CORD test)}} \\
\rzero{} / \rone{} & .456 / .730 & .912 / .881 \\
\rtwo{} / \rthree{} & .302 / \textbf{.746} & .768 / .770 \\
\rfour{} & .726 & \textbf{.939} \\
\bottomrule
\end{tabular}
\caption{Robustness checks: three CORD retraining seeds, the dev split, and the 4B
backbone. Bold marks the best value per block and column among the regimes shown, seed 0 only; at 4B \rthree{} leads on point ($.746$ against \rfour{}'s $.726$, $-2.0$ $[-4.5,+0.4]$).}
\label{tab:robust}
\end{table}

\paragraph{Robustness.} Intervals that also resample training seeds change no headline
conclusion (Appendix~\ref{app:tables}; the CORD line gain becomes $+6.0$ $[2.5,9.4]$). Seeds and the dev split hold the ordering fixed, \rfour{} leading both sides on dev (Tables~\ref{tab:robust} and~\ref{tab:wr}); arms outside those seed blocks are single runs.

\begin{table}[t]
\centering\footnotesize\setlength{\tabcolsep}{3pt}
\begin{tabular}{lcc}
\toprule
WildReceipt (n=100) & Point F1 & Line Acc \\
\midrule
\rzero{} & .414 & .835 \\
\rone{} & .453/.441/.379 & .785/.780/.775 \\
\rtwo{} & .387/.398/.387 & \textbf{.903}/.890/.878 \\
\rthree{} & .527/.535/.506 & .878/.878/.843 \\
\rthreeb{} & .423 & .878 \\
\rfour{} & \textbf{.591}/.558/.571 & \textbf{.890}/.878/.880 \\
\rpipe{} & .399 & \textbf{.905} \\
\rfoursh{} & \textbf{.573} & .715 \\
\rlndir{} & .521 & .858 \\
\rptdir{} & \textbf{.596} & .858 \\
\rfourn{} & \textbf{.610} & \textbf{.888} \\
\bottomrule
\end{tabular}
\caption{WildReceipt replication, seeds 0/1/2 (seed 0 is the pre-registered run and
carries the CIs; \rtwo{}, \rthreeb{} and \rpipe{} are post-hoc additions). Bold marks the best
value in each column and every value within its paired interval, on seed 0; the other two seeds
carry no interval and are never bolded. On line \rpipe{} ($.905$) and \rtwo{} ($.903$) lead
\rfour{} and on point \rfourn{} ($.610$) does, all three within the interval, which is the
finding of Section~\ref{sec:results}. \rfour{} vs
\rthree{}: point $+6.4$ $[1.8,11.1]$, line $+1.2$ $[-1.5,4.0]$; \rfour{} vs \rzero{}
line $+5.5$ $[1.8,9.2]$; \rone{} line damage replicates ($-5.0$).}
\label{tab:wr}
\end{table}

\paragraph{Two standard alternatives do not produce the gain.} Two familiar arrangements fail in different ways (Appendix~\ref{app:pipe}). A two-stage pipeline leads the CORD line column, but only at twice the inference cost, and it collapses where extraction is hardest (WildReceipt point $.399$, $19$ below \rfour{}). Sequential transfer \citep{phang2018sentence} costs nothing extra and removes the cross-task damage, holding point at $.801$ against \rtwo{}'s $.480$, but gains nothing on line: both orders land below zero-shot while \rfour{} exceeds them by $+5.1$ and $+5.3$. The damage is an ordering artefact; the gain is not.

\paragraph{Replication on a second dataset and backbone.} Under a pre-registered protocol (Appendix~\ref{app:prereg}) we rebuilt the benchmark on
WildReceipt, English receipts under a different twelve-category schema, with the CORD recipe
unchanged. \rfour{} leads every pre-registered regime on point ($+6.4$ over \rthree{}, $[1.8,11.1]$) and
sits within the paired interval of the best line regime, while two post-hoc arms are nominally higher on point with unresolvable differences (Table~\ref{tab:wr}).  Unlike on CORD, plain mixing buys the point side ($+7.4$ over an unstable \rone{}, $+2.0$ once both models must parse) but pays $2.5$ line points against
\rtwo{} ($[-4.8,-0.5]$), which does not self-damage here: mixing trades one granularity for the other. Scored on the full 467-image test split, which needs no new annotation on the fine side, the fine-side contrasts hold and tighten (Appendix~\ref{app:train}). At 4B (Appendix~\ref{app:4b}) the coarse-side pattern amplifies, \rtwo{} and \rthree{} falling $14$ points below zero-shot on line with \rfour{} alone above it. It is also the one configuration of six in which mixed training clears both matched single-task models, by $+1.5$ point and $+0.3$ line, neither resolvable and both over damaged baselines.

Which side gains flips across datasets. What holds at the shared recipe is the first half of the dominance pattern: no alternative regime measurably beats \rfour{} on either side, the one exception in the study being mixed training on FUNSD's fine side at the lower rate ($+9.1$, $[2.2,16.3]$). The converse is weaker: every \emph{pre-registered} alternative falls measurably behind on at least one side, but four post-hoc arm-corpus pairs fall behind on neither, \rpipe{} and \rlndir{} on CORD, \rfourn{} on WildReceipt and \rlndir{} again on FUNSD, all eight intervals including zero (Appendix~\ref{app:tables}).

\begin{table}[t]
\centering\footnotesize\setlength{\tabcolsep}{3pt}
\begin{tabular}{lccc}
\toprule
FUNSD (n=50) & Point F1 & Line Acc & \texttt{doc\_type} \\
\midrule
\rzero{} & .117 & \textbf{.915} & .900 \\
\rone{} & .358 & \textbf{.905} & \textbf{.940} \\
\rtwo{} & .232 & .810 & .560 \\
\rthree{} & \textbf{.426} & .770 & .520 \\
\rfour{} & \textbf{.430} & \textbf{.920} & \textbf{1.000} \\
\rlndir{} & \textbf{.431} & \textbf{.935} & \textbf{1.000} \\
\midrule
\rptdir{} & .298 & .810 & .520 \\
\rfourn{} & \textbf{.402} & .750 & .520 \\
\rfoursh{} & \textbf{.408} & .650 & .520 \\
\bottomrule
\end{tabular}
\caption{Third domain: scanned business forms, official split, seed 0 at the shared recipe; seed replications and a learning-rate and epoch sweep are in Table~\ref{tab:fssweep}, and \rthreeb{} ($.421/.875/.820$) is omitted for space. The majority \texttt{doc\_type} is $.520$, and at this recipe \rthree{} and both content-free controls assign it to all $50$ documents in every seed, while \rfour{} recovers the gold distribution. Bold marks the best value in each column and
every value within its paired interval; on line that is four of the eight arms, so the coarse
side separates the collapsed arms from the rest and not \rfour{} from the untuned model.}
\label{tab:funsd}
\end{table}

\paragraph{A third genre, and a collapse that conditioning prevents.} Both corpora above are receipts, so we repeated the study on FUNSD, $199$ scanned 1990s business documents whose gold entities give the fine task and whose four coarse facets the same committee annotated (Appendix~\ref{app:funsd}). What it adds is not a larger gain but a failure conditioning prevents (Table~\ref{tab:funsd}). Against the untuned model \rfour{}'s coarse accuracy is $+0.5$, one facet decision in $200$: no aggregate coarse-side gain. What differs is the baselines: under the shared recipe only conditioning and its fields-to-facets half reach zero-shot on the coarse side, and on the semantic facet \rthree{} assigns the majority \texttt{doc\_type} to all $50$ documents where zero-shot reaches $.900$, while \rfour{} recovers the gold distribution exactly. Three seeds hold both sides fixed, with \texttt{doc\_type} $1.000/.960/.980$ against $.520$ every time (Table~\ref{tab:fssweep}). The neutral control collapses identically ($+17.0$, $[12.0,22.5]$): the conditioning \emph{content}, not the filled slot, protects the facet.

Two controls change how to read it. It is not a task-proportion artefact: oversampling the coarse task two and four times leaves \texttt{doc\_type} at the majority label ($+0.0$). It is a learning-rate effect: at $2{\times}10^{-5}$ mixed training recovers to $.890$ line and $.880$ \texttt{doc\_type}, neither resolvably below zero-shot, while four epochs makes it worse ($.700$). Conditioning is unmoved by the rate ($.910$) and beats the best-tuned mixed arm on the semantic facet at either rate, $+12.0$ ($[4.0,22.0]$) from its own recipe and $+8.0$ ($[2.0,16.0]$) at the lower one; the macro margins, $+3.0$ and $+2.0$, resolve only at the first. At the lower rate no regime reinforces on FUNSD against single-task models retrained there, and on CORD conditioning trades rather than reinforces ($-1.4$ point, $+3.8$ line): mixed training's failure survives the change on both corpora and conditioning's repair does not. Two pre-registered outcomes also went against us (Appendix~\ref{app:funsd}).

\section{What the Analysis Battery Shows}
\label{sec:analysis}

Four instruments, each with its own control, reported as descriptive. Three run in a prompt format only the conditioned arms saw in training; only the probes run on plain prompts (Appendices~\ref{app:tables} and~\ref{app:attrib}).

\paragraph{Encoding.} Probes on one task's plain prompt, with a random-label control task
\citep{hewitt2019designing}, recover the other task's labels under every regime (selectivity $.36$--$.46$ at the cross-validated layer on three facets; the fourth is skewed and uses balanced
accuracy). \rfour{} adds nothing over zero-shot that we can resolve, $+2.0$ $[-4.0,+8.1]$ on
\texttt{store\_type}, so an injection account is unnecessary rather than impossible; \rthree{} probes $9.1$ below it there ($[1.0,17.2]$). On two other facets \rfour{}
itself decodes below zero-shot.

\paragraph{Use.} Supplying the point task with a gold or corrupted hypothesis (Table~\ref{tab:interv}) makes three of six fine-tuned regimes sensitive while zero-shot is not. So does \rfourn{}, more strongly than \rfour{} ($+2.2$ $[1.1,3.3]$ against $+1.5$ $[0.4,2.6]$), though its slots asserted nothing in training: the instrument measures format exposure, not learned use of cross-granularity content. Against each model's own plain prompt the two dissociate. \rfourn{} gains from a gold hypothesis ($+1.1$, $[0.2,2.1]$); \rfour{} does not ($-0.4$, $[-1.6,+0.8]$) and is harmed by a corrupted one ($-1.9$, $[-3.0,-0.7]$). That is what internalization predicts, an external hypothesis being redundant for a model that already encodes it, but distractibility predicts the harm as well and these instruments cannot separate them.

\paragraph{Attribution and modality.} Gradient attribution concentrates on the annotated
evidence above a count-matched null for every model, and on the two facets that survive the
lexical null \rfour{} is highest (\texttt{store\_type} $1.88\!\to\!2.47$, paired median
difference over zero-shot $+0.59$ $[0.30,0.78]$). It is not separable from \rthree{} there, but it exceeds the neutral control on both facets ($+0.139$ and $+0.222$ paired median lift, both excluding zero), so unlike the intervention this instrument is not reproduced by format alone. \rfour{} is also the only regime reading the
facets better from the field list than from the image alone ($+1.0$ $[-1.5,+3.8]$, all others
negative): a reversal of which channel dominates, not a broken route, unresolvable against
zero-shot ($+2.8$ $[-1.0,+6.3]$) and with format familiarity predicting its sign.

\paragraph{What this does and does not support.} The instruments bound injection below their own resolution but do not establish a readout mechanism: patching was inconclusive (Appendix~\ref{app:patch}) and the format confound is bounded for two of the three that share it, in opposite directions. The readout reading is the one we favour, not a result.

\section{Conclusion}

Mutual reinforcement does not come free with joint training: across three corpora mixed co-training clears both single-task models nowhere at the main scale, while conditioning does on two and trades on the third.

\section*{Limitations}

Our evidence covers two receipt corpora and one corpus of scanned business forms, one VLM
family at two scales and LoRA fine-tuning; full fine-tuning and further genres are left open,
and the FUNSD corpus has $149$ training and $50$ test documents, so its intervals are correspondingly wide, with three seeds run only for mixed, conditioned and the neutral control. The mechanism claim
is deliberately scoped: three of the four analysis instruments share a prompt-format confound
with the effect they measure, so we report the battery as descriptive and leave a
representation-level account to future work. Line labels come from an LLM committee, and blind human re-annotation bounds their noise at roughly 4\% on CORD, 8\% on WildReceipt and 9\% on FUNSD; because
every regime comparison is paired on identical labels, that noise shifts absolute levels rather
than the reported contrasts. Test sets hold 50--100 documents, so per-facet and cross-regime contrasts are reported as point estimates and only the contrasts listed in Appendix~\ref{app:tables} carry intervals. Finally, conditioned training presumes the other
granularity's gold labels exist at training time; where they must first be produced, as ours
were, the cost of the recipe includes that annotation step.

\section*{Ethics Statement}

All three source corpora are public research datasets used under their research terms, and we
redistribute no images: Doc-MRE ships as an annotation layer keyed to the original file
identifiers. Human validation was carried out by three project-affiliated annotators working independently from a written guide per facet schema, the receipt one reproduced in Appendix~\ref{app:guide}, with no crowd workers employed.

\bibliography{custom}

\begin{thebibliography}{21}
\providecommand{\natexlab}[1]{#1}

\bibitem[{Alain and Bengio(2016)}]{alain2016understanding}
Guillaume Alain and Yoshua Bengio. 2016.
\newblock Understanding intermediate layers using linear classifier probes.
\newblock \emph{arXiv preprint arXiv:1610.01644}.

\bibitem[{Bai et~al.(2025)Bai, Cai, Chen, Chen, Chen, Cheng, Deng, Ding, Gao,
  Ge et~al.}]{bai2025qwen3}
Shuai Bai, Yuxuan Cai, Ruizhe Chen, Keqin Chen, Xionghui Chen, Zesen Cheng,
  Lianghao Deng, Wei Ding, Chang Gao, Chunjiang Ge, et~al. 2025.
\newblock Qwen3-vl technical report.
\newblock \emph{arXiv preprint arXiv:2511.21631}.

\bibitem[{Caruana(1997)}]{caruana1997multitask}
Rich Caruana. 1997.
\newblock Multitask learning.
\newblock \emph{Machine learning}, 28(1):41--75.

\bibitem[{Elazar et~al.(2021)Elazar, Ravfogel, Jacovi, and
  Goldberg}]{elazar2021amnesic}
Yanai Elazar, Shauli Ravfogel, Alon Jacovi, and Yoav Goldberg. 2021.
\newblock Amnesic probing: Behavioral explanation with amnesic counterfactuals.
\newblock \emph{Transactions of the Association for Computational Linguistics},
  9:160--175.

\bibitem[{Gan et~al.(2025)Gan, Cai, Wei, Liang, Ni, and Mori}]{gan2025m}
Chengguang Gan, Zhixi Cai, Yanbin Wei, Yunhao Liang, Shiwen Ni, and Tatsunori
  Mori. 2025.
\newblock {M-MRE}: Extending the mutual reinforcement effect to multimodal
  information extraction.
\newblock \emph{arXiv preprint arXiv:2504.17353}.

\bibitem[{Hewitt and Liang(2019)}]{hewitt2019designing}
John Hewitt and Percy Liang. 2019.
\newblock Designing and interpreting probes with control tasks.
\newblock In \emph{Proceedings of the 2019 conference on empirical methods in
  natural language processing and the 9th international joint conference on
  natural language processing (emnlp-ijcnlp)}, pages 2733--2743.

\bibitem[{Hsieh et~al.(2023)Hsieh, Li, Yeh, Nakhost, Fujii, Ratner, Krishna,
  Lee, and Pfister}]{hsieh2023distilling}
Cheng-Yu Hsieh, Chun-Liang Li, Chih-Kuan Yeh, Hootan Nakhost, Yasuhisa Fujii,
  Alex Ratner, Ranjay Krishna, Chen-Yu Lee, and Tomas Pfister. 2023.
\newblock Distilling step-by-step! outperforming larger language models with
  less training data and smaller model sizes.
\newblock In \emph{Findings of the Association for Computational Linguistics:
  ACL 2023}, pages 8003--8017.

\bibitem[{Hu et~al.(2022)Hu, Shen, Wallis, Allen-Zhu, Li, Wang, Wang, and
  Chen}]{hu2022lora}
Edward~J Hu, Yelong Shen, Phillip Wallis, Zeyuan Allen-Zhu, Yuanzhi Li, Shean
  Wang, Liang Wang, and Weizhu Chen. 2022.
\newblock {LoRA}: Low-rank adaptation of large language models.
\newblock In \emph{International Conference on Learning Representations}.

\bibitem[{Huang et~al.(2022)Huang, Lv, Cui, Lu, and Wei}]{huang2022layoutlmv3}
Yupan Huang, Tengchao Lv, Lei Cui, Yutong Lu, and Furu Wei. 2022.
\newblock Layoutlmv3: Pre-training for document ai with unified text and image
  masking.
\newblock In \emph{Proceedings of the 30th ACM international conference on
  multimedia}, pages 4083--4091.

\bibitem[{Jaume et~al.(2019)Jaume, Ekenel, and Thiran}]{jaume2019funsd}
Guillaume Jaume, Hazim~Kemal Ekenel, and Jean-Philippe Thiran. 2019.
\newblock {FUNSD}: A dataset for form understanding in noisy scanned documents.
\newblock In \emph{2019 International Conference on Document Analysis and
  Recognition Workshops (ICDARW)}, volume~2, pages 1--6. IEEE.

\bibitem[{Kim et~al.(2022)Kim, Hong, Yim, Nam, Park, Yim, Hwang, Yun, Han, and
  Park}]{kim2022ocr}
Geewook Kim, Teakgyu Hong, Moonbin Yim, JeongYeon Nam, Jinyoung Park, Jinyeong
  Yim, Wonseok Hwang, Sangdoo Yun, Dongyoon Han, and Seunghyun Park. 2022.
\newblock Ocr-free document understanding transformer.
\newblock In \emph{European Conference on Computer Vision}, pages 498--517.
  Springer.

\bibitem[{Lopez-Paz et~al.(2015)Lopez-Paz, Bottou, Sch{\"o}lkopf, and
  Vapnik}]{lopez2015unifying}
David Lopez-Paz, L{\'e}on Bottou, Bernhard Sch{\"o}lkopf, and Vladimir Vapnik.
  2015.
\newblock Unifying distillation and privileged information.
\newblock \emph{arXiv preprint arXiv:1511.03643}.

\bibitem[{Park et~al.(2019)Park, Shin, Lee, Lee, Surh, Seo, and
  Lee}]{park2019cord}
Seunghyun Park, Seung Shin, Bado Lee, Junyeop Lee, Jaeheung Surh, Minjoon Seo,
  and Hwalsuk Lee. 2019.
\newblock {CORD}: a consolidated receipt dataset for post-{OCR} parsing.
\newblock In \emph{Workshop on document intelligence at NeurIPS}, volume 2019,
  page~5.

\bibitem[{Phang et~al.(2018)Phang, F{\'e}vry, and Bowman}]{phang2018sentence}
Jason Phang, Thibault F{\'e}vry, and Samuel~R Bowman. 2018.
\newblock Sentence encoders on stilts: Supplementary training on intermediate
  labeled-data tasks.
\newblock \emph{arXiv preprint arXiv:1811.01088}.

\bibitem[{Simonyan et~al.(2013)Simonyan, Vedaldi, and
  Zisserman}]{simonyan2013deep}
Karen Simonyan, Andrea Vedaldi, and Andrew Zisserman. 2013.
\newblock Deep inside convolutional networks: Visualising image classification
  models and saliency maps.
\newblock \emph{arXiv preprint arXiv:1312.6034}.

\bibitem[{Snell et~al.(2022)Snell, Klein, and Zhong}]{snell2022learning}
Charlie Snell, Dan Klein, and Ruiqi Zhong. 2022.
\newblock Learning by distilling context.
\newblock \emph{arXiv preprint arXiv:2209.15189}.

\bibitem[{Standley et~al.(2020)Standley, Zamir, Chen, Guibas, Malik, and
  Savarese}]{standley2020tasks}
Trevor Standley, Amir Zamir, Dawn Chen, Leonidas Guibas, Jitendra Malik, and
  Silvio Savarese. 2020.
\newblock Which tasks should be learned together in multi-task learning?
\newblock In \emph{International conference on machine learning}, pages
  9120--9132. PMLR.

\bibitem[{Sun et~al.(2021)Sun, Kuang, Yue, Lin, and Zhang}]{sun2021spatial}
Hongbin Sun, Zhanghui Kuang, Xiaoyu Yue, Chenhao Lin, and Wayne Zhang. 2021.
\newblock Spatial dual-modality graph reasoning for key information extraction.
\newblock \emph{arXiv preprint arXiv:2103.14470}.

\bibitem[{Vapnik and Vashist(2009)}]{vapnik2009new}
Vladimir Vapnik and Akshay Vashist. 2009.
\newblock A new learning paradigm: Learning using privileged information.
\newblock \emph{Neural networks}, 22(5-6):544--557.

\bibitem[{Yu et~al.(2020)Yu, Kumar, Gupta, Levine, Hausman, and
  Finn}]{yu2020gradient}
Tianhe Yu, Saurabh Kumar, Abhishek Gupta, Sergey Levine, Karol Hausman, and
  Chelsea Finn. 2020.
\newblock Gradient surgery for multi-task learning.
\newblock \emph{Advances in neural information processing systems},
  33:5824--5836.

\bibitem[{Zheng et~al.(2023)Zheng, Chiang, Sheng, Zhuang, Wu, Zhuang, Lin, Li,
  Li, Xing et~al.}]{zheng2023judging}
Lianmin Zheng, Wei-Lin Chiang, Ying Sheng, Siyuan Zhuang, Zhanghao Wu, Yonghao
  Zhuang, Zi~Lin, Zhuohan Li, Dacheng Li, Eric Xing, et~al. 2023.
\newblock Judging llm-as-a-judge with mt-bench and chatbot arena.
\newblock \emph{Advances in neural information processing systems},
  36:46595--46623.

\end{thebibliography}

\appendix

% Artifacts appendix, commented out 2026-07-26 at the author's request.
% Restore for camera-ready and replace \anonurl with the real repository URL.
% \section{Artifacts}
% \label{app:artifacts}
%
% The three annotation layers (line labels, evidence links, and the FUNSD entity decoding), the training and evaluation code for every regime and control, the three-judge committee prompts, the human annotation interface, and the per-sample prediction files behind every interval in this paper are released at \anonurl{} under the source corpora's licences. The pre-registrations and the frozen judge prompts are included verbatim so the reported gates can be rechecked against the shipped labels.

\section{Training and Evaluation Details}
\label{app:train}

All runs use Qwen3-VL-8B-Instruct (Appendix~\ref{app:4b}: 4B) with LoRA rank 64, $\alpha$ 128, which makes $174.6$M of the $8.94$B parameters trainable ($1.95\%$), dropout $0.05$ on all attention and MLP projections, bf16, learning rate $10^{-4}$ with a
one-cycle schedule, batch size 1 with gradient accumulation 8, two epochs, loss on answer
tokens only. \rone{}/\rtwo{} see 794 examples, \rthree{}/\rfour{}/\rfoursh{} 1{,}588 (equal
per-task supervision); the step-matched control trains \rone{}/\rtwo{} four epochs: point
$.803$ vs $.804$, line $.894$ vs $.899$, both within noise. A learning-rate control retrains every CORD arm at $2{\times}10^{-5}$. \rtwo{} reaches line $.894$, still below zero-shot, so the self-damage of line-only tuning is not a recipe artifact; \rone{} reaches point $.775$ and \rthree{} $.743$ point and $.891$ line, leaving mixed training below both matched single-task models again ($-3.3$ $[-7.3,+0.6]$ fine, $-0.3$ $[-1.5,+1.0]$ coarse) exactly as at $10^{-4}$ ($-2.6$, $-0.5$). Conditioning at $10^{-4}$ leads the best-tuned mixed arm on line by $+5.6$ $[3.0,8.1]$. Retrained at $2{\times}10^{-5}$ itself it reaches point $.761$ and line $.932$, which against the single-task arms at that rate is $-1.4$ $[-5.3,+2.6]$ on the fine side, $-2.5$ $[-5.0,-0.1]$ once both models must parse, and $+3.8$ $[1.0,6.6]$ on the coarse side. It therefore trades at the lower rate where it reinforces at the shared one, the same pattern as on FUNSD, so the reinforcement result is a property of the recipe on both corpora while the failure of mixed training is not. Evaluation is greedy decoding with plain prompts;
WildReceipt (300 train, 100 test) and FUNSD (149 train, 50 test) reuse the CORD recipe with no per-corpus tuning, per the pre-registration; The fine side also scales to WildReceipt's whole official test split, 467 images after the same filter, which needs no new coarse annotation: point F1 is .592 for \rfour{}, .536 for \rthree{}, .435 for \rone{} and .398 for \rtwo{}, and the contrasts the paper rests on hold with intervals roughly half as wide: \rfour{}$-$\rthree{} $+5.6$ $[3.5,7.6]$, \rfour{}$-$\rone{} $+15.6$ $[12.8,18.4]$ and \rthree{}$-$\rone{} $+10.1$ $[7.7,12.5]$. Parse rates on that split are 339/467 for \rone{}, 418/467 for \rtwo{}, 401/467 for \rthree{} and 414/467 for \rfour{}, and the same correction applied above gives $+4.7$ $[3.4,5.9]$ on 383 samples, $+7.3$ $[5.8,8.8]$ on 322 and $+2.8$ $[1.8,3.8]$ on 325. The first 100 indices of that split are the set reported in Table~\ref{tab:wr}, byte for byte, and reproduce its numbers exactly. The WildReceipt seed replications, the \rtwo{}, \rthreeb{} and \rpipe{} regimes, and
the shuffled and neutral-content controls were added after the pre-registered run and are
reported as post-hoc robustness checks.

All experiments ran on a single server with eight RTX PRO 6000 GPUs (96\,GB each); every
training or evaluation job fits on one GPU, and jobs were parallelized across the cards. One regime trains in 20--60 GPU-minutes; the complete study, all regimes, seeds, backbones, corpora, the recipe controls and the analysis battery, totals roughly 160 GPU-hours. LLM-committee annotation for all three corpora cost under \$40 of API usage.

\section{Additional Result Detail}
\label{app:tables}

\paragraph{Headline intervals.} CORD test, paired bootstrap, 10k resamples:
\rfour{}$-$\rtwo{} line $+4.8$ $[2.3,7.3]$; \rfour{}$-$\rthree{} line $+5.3$ $[2.5,8.1]$;
\rfour{}$-$\rzero{} line $+3.5$ $[1.0,6.1]$; \rfour{}$-$\rone{} point $+0.5$ $[-1.8,+2.6]$;
\rfour{}$-$\rthree{} point $+3.0$ $[1.1,5.2]$; \rfour{}$-$\rthreeb{} point $+1.3$
$[-1.1,+3.7]$; \rone{}$-$\rzero{} line $-5.3$ $[-8.8,-2.0]$; \rtwo{}$-$\rzero{} point $-7.5$
$[-11.0,-4.4]$. The largest contrasts (both damage effects, the CORD line gains, WR point
over \rone{}) sit far from their interval boundaries; the smaller ones ($+3.0$, $+3.5$,
$+6.4$) sit closer, and we treat them as supporting rather than load-bearing.

\begin{table}[t]
\centering\footnotesize\setlength{\tabcolsep}{2pt}
\begin{tabular}{lcccc}
\toprule
Facet & bf16 & fp32 & overlap & non-lex.\ lift \\
\midrule
store\_type & 1.88/2.47 & 1.94/2.47 & 9\% & 1.90/2.45 \\
has\_surcharge & 1.10/1.55 & 1.17/1.54 & 5\% & 1.17/1.54 \\
has\_discount & 1.31/1.51 & 1.41/1.87 & 14\% & 1.41/1.87 \\
payment & 1.16/1.22 & 1.16/1.31 & 68\% & 1.00/0.99 \\
\bottomrule
\end{tabular}
\caption{Attribution validity checks (median lift, zero-shot/\rfour{}; all four regimes
in Figure~\ref{fig:attrib}). A float32 recomputation moves the two surviving facets by at most $.07$ and
\texttt{has\_discount}, whose $n$ is $12$, by $.36$. ``Overlap'' is the fraction of evidence fields whose
text lexically overlaps the answer; ``non-lex.\ lift'' recomputes lift on non-overlapping
evidence only. \texttt{payment\_method} fails the null and is excluded.}
\label{tab:attribval}
\end{table}

\paragraph{Pipeline contrasts.} \rfour{}$-$\rpipe{} line: CORD $-1.5$ $[-3.5,+0.3]$,
WildReceipt $-1.5$ $[-4.5,+1.5]$; \rfour{}$-$\rpipe{} point: CORD $+0.4$ $[-2.0,+2.6]$,
WildReceipt $+19.2$ $[13.6,25.0]$. \rpipe{}, added post hoc like the other arms in this paragraph, leads \rfour{} on line; \rtwo{} on
WildReceipt line, \rthree{} on 4B point, and \rfourn{} and \rptdir{} on WildReceipt point
also lead it nominally. Every one of these intervals includes zero.

\paragraph{Parse-rate correction of the fine-side gains.} The fine-side damage claim is reported on the subset both models parse (Section~\ref{sec:results}); the same correction belongs on the gains wherever the arms differ in parse rate. On CORD they barely do, every trained arm parsing $99$ of $99$, so every CORD gain is identical corrected and uncorrected and that corpus carries no format component; the only CORD figure the correction moves is the line-only damage, $-7.6$ to $-5.5$, which Section~\ref{sec:results} already reports. On WildReceipt the arms differ by more than the gains are large ($.930$ zero-shot, $.750$ \rone{}, $.860$ \rthree{}, $.880$ \rfour{}). Restricting each contrast to samples both arms parsed: \rfour{}$-$\rthree{} $+6.4$ $[1.8,11.1]$ becomes $+6.1$ $[3.2,9.1]$ on $81$ samples, \rfour{}$-$\rone{} $+13.8$ $[7.6,19.9]$ becomes $+8.7$ $[5.4,12.1]$ on $70$, \rthree{}$-$\rone{} $+7.4$ $[2.7,12.1]$ becomes $+2.0$ $[0.4,3.6]$ on $72$, and \rone{}$-$\rzero{} $+3.9$ $[-2.7,+10.7]$ becomes $+15.8$ $[11.1,20.5]$ on $69$, that last one having been masked by \rone{}'s own parse failures. Every contrast keeps its sign and every one that was resolvable stays resolvable, so no verdict changes; the headline moves by $0.3$ points and two secondary numbers shrink by a third and by three quarters. The one contrast whose sign does move is \rfour{} against \rfourn{}, $-1.9$ $[-7.0,+2.8]$ raw and $+1.2$ $[-0.5,+3.1]$ corrected, and since neither interval excludes zero the two arms are indistinguishable under either reading, which is what the claim about the neutral slot reproducing that gain rests on. On FUNSD the arms differ too ($34/50$ \rone{}, $41/50$ \rthree{}, $43/50$ \rfour{}, $40/50$ \rfourn{}) and the corrected figures are \rfour{}$-$\rone{} $+7.2$ $[-0.7,+14.8]$ to $+1.9$ $[-1.5,+5.8]$ on $31$ samples, \rfour{}$-$\rthree{} $+0.4$ $[-3.3,+4.2]$ to $-1.3$ $[-4.2,+1.2]$ on $40$, and \rthree{}$-$\rone{} $+6.8$ $[-0.5,+14.2]$ to $+1.7$ $[-1.5,+5.2]$ on $32$. None of those was resolvable before the correction and none is after, which is why the paper reports FUNSD's fine side as directional throughout; the corrected \rfour{}$-$\rone{} figure is still positive, so the reinforcement verdict on that corpus does not turn on the correction.

\paragraph{Human-label rescoring (WildReceipt).} Scoring every regime against the
three-annotator human consensus instead of the committee majority lowers line accuracy by $2.3$ to
$4.5$ points everywhere: \rzero{} $.835\!\to\!.803$, \rone{} $.785\!\to\!.740$,
\rtwo{} $.903\!\to\!.880$, \rthree{} $.878\!\to\!.840$, \rthreeb{}
$.878\!\to\!.840$, \rfour{} $.890\!\to\!.853$, \rpipe{} $.905\!\to\!.873$,
\rfoursh{} $.715\!\to\!.690$, \rlndir{} $.858\!\to\!.815$, \rptdir{} $.858\!\to\!.820$ and \rfourn{} $.887\!\to\!.855$. Two orderings change: \rpipe{} is above \rtwo{} under the committee labels and below it under the human ones, and \rfour{} falls $0.2$ points below \rfourn{}, which the committee labels put $0.3$ the other way. The contrasts \rfour{} depends on
widen rather than narrow, \rfour{}$-$\rtwo{} from $-1.3$ $[-4.5,+1.8]$ to $-2.8$
$[-6.2,+0.8]$ and \rfour{}$-$\rpipe{} from $-1.5$ $[-4.5,+1.5]$ to $-2.0$ $[-5.0,+1.0]$, so
the no-loss claim on that side is weaker against human labels while remaining unresolvable.

\begin{table}[t]
\centering\footnotesize\setlength{\tabcolsep}{2pt}
\begin{tabular}{lccc}
\toprule
Probe target & \rzero{} & \rfour{} & others \\
\midrule
store\_type & .44 & .46 & .41/.44/.38 \\
payment\_method & .44 & .39 & .41/.36/.37 \\
has\_surcharge & .45 & .44 & .43/.45/.46 \\
\bottomrule
\end{tabular}
\caption{Probe selectivity (accuracy minus a random-label control probe, both at the
layer chosen by train-side cross-validation) for line facets decoded from point-task states;
the last column lists \rone{}/\rtwo{}/\rthree{} in that order. Selecting the layer on the test
side instead inflates individual cells by up to $.10$, so the cross-validated layer is what we
report. All regimes clear the control by a wide margin on all three facets and the spread across regimes, $.36$ to $.46$, is smaller than that inflation. Selectivity is uninformative for \texttt{has\_discount}, whose
marginal-preserving control scores $.873$ against an $.899$ majority; balanced accuracy is
reported for it instead (Appendix~\ref{app:tables}).}
\label{tab:selectivity}
\end{table}

\begin{table*}[t]
\centering\small
\begin{tabular}{lccc|ccc}
\toprule
& \multicolumn{3}{c|}{point F1 under} & \multicolumn{3}{c}{paired contrast (CI)} \\
Regime & plain & gold & corrupt & gold $-$ corrupt & gold $-$ plain & corrupt $-$ plain \\
\midrule
\rzero{} & .555 & .558 & .564 & $-0.7$ $[-3.0,+1.8]$ & $+0.2$ $[-2.6,+3.1]$ & $+0.9$ $[-2.2,+3.8]$ \\
\rone{} & .804 & .814 & .795 & $+1.9$ $[0.7,3.3]$ & $+1.0$ $[-0.7,+2.8]$ & $-0.9$ $[-2.8,+0.8]$ \\
\rtwo{} & .480 & .488 & .479 & $+0.9$ $[-2.5,+4.3]$ & $+0.8$ $[-2.8,+4.3]$ & $-0.1$ $[-4.3,+3.9]$ \\
\rthree{} & .778 & .789 & .770 & $+2.0$ $[0.5,3.8]$ & $+1.1$ $[-0.5,+3.2]$ & $-0.8$ $[-2.6,+1.0]$ \\
\rthreeb{} & .795 & .803 & .792 & $+1.1$ $[-0.8,+3.4]$ & $+0.8$ $[-0.9,+2.7]$ & $-0.3$ $[-2.7,+1.8]$ \\
\rfour{} & .808 & .805 & .790 & $+1.5$ $[0.4,2.6]$ & $-0.4$ $[-1.6,+0.8]$ & $\mathbf{-1.9}$ $[-3.0,-0.7]$ \\
\rfoursh{} & .776 & .773 & .773 & $+0.0$ $[-0.4,+0.5]$ & $-0.2$ $[-1.3,+0.8]$ & $-0.3$ $[-1.4,+0.8]$ \\
\rfourn{} & .787 & .798 & .776 & $\mathbf{+2.2}$ $[1.1,3.3]$ & $\mathbf{+1.1}$ $[0.2,2.1]$ & $-1.1$ $[-2.4,+0.2]$ \\
\bottomrule
\end{tabular}
\caption{Point-side input interventions (point F1), all eight regimes. Sensitivity is
F1(gold) $-$ F1(corrupted), paired per sample. Three of the six fine-tuned regimes are
individually sensitive; \rtwo{} and \rthreeb{} are not, and \rfoursh{}, trained on
systematically wrong context, has learned to ignore the slot entirely ($+0.0$). Read against each model's own plain prompt only two intervals are resolvable, both bolded: \rfour{} is harmed by a corrupted hypothesis and \rfourn{}, which saw the format without the content, is helped by a gold one. The sensitivity column additionally excludes zero for \rone{}, \rthree{} and \rfourn{}.
Direct between-regime contrasts stay within noise (Appendix~\ref{app:tables}).}
\label{tab:interv}
\end{table*}

\paragraph{Direction arms.} \rlndir{} and \rptdir{} are built by recombining rows already
present in the \rfour{} and \rthree{} training files, so no prompt is regenerated:
\rlndir{} takes \rfour{}'s field-conditioned line examples plus \rthree{}'s plain point
examples, \rptdir{} takes \rfour{}'s hypothesis-conditioned point examples plus \rthree{}'s
plain line examples. Both keep 1{,}588 examples, the same images, the same targets and equal
per-task supervision, so the only difference from \rfour{} is that one direction's context
slot is removed; a per-image assertion checks that the conditioned and plain rows of a task
carry identical targets. CORD: \rlndir{} point $.793$, line $.944$; \rptdir{} point $.775$,
line $.891$. Contrasts: \rlndir{}$-$\rthree{} line $+5.1$ $[2.3,8.1]$;
\rlndir{}$-$\rfour{} line $-0.3$ $[-1.8,+1.5]$; \rptdir{}$-$\rthree{} point $-0.4$
$[-2.5,+1.8]$; \rfour{}$-$\rlndir{} point $+1.5$ $[0.0,+3.0]$. WildReceipt, where the roles
swap: \rptdir{} point $.596$, line $.858$; \rlndir{} point $.521$, line $.858$;
\rptdir{}$-$\rthree{} point $+7.0$ $[3.2,10.9]$; \rfour{}$-$\rptdir{} point $-0.5$
$[-5.0,+3.8]$; \rlndir{}$-$\rthree{} line $-2.0$ $[-5.2,+1.0]$; \rfour{}$-$\rlndir{} line
$+3.3$ $[1.0,5.5]$; \rptdir{}$-$\rfourn{} point $-1.4$ $[-4.5,+1.2]$.

\paragraph{Sequential transfer.} Each arm continues training an existing single-task adapter
on the other task's data with the recipe otherwise unchanged, which is the STILTs arrangement
\citep{phang2018sentence} applied to our pair. CORD: point after line ($\rseq{}$
ln$\to$pt) point $.804$, line $.894$; line after point ($\rseq{}$ pt$\to$ln) point $.801$,
line $.897$. Contrasts against the single-task models: ln$\to$pt point $+0.1$ $[-2.4,+2.5]$
and line $+3.5$ $[0.0,+7.3]$ against \rone{}; pt$\to$ln point $+32.2$ $[27.5,36.9]$ against
\rtwo{} and line $-0.3$ $[-2.0,+1.8]$ against \rtwo{}. Against \rfour{}: line $-5.3$
$[-8.3,-2.5]$ and $-5.1$ $[-7.8,-2.3]$ for the two orders, point $-0.7$ $[-2.6,+1.3]$ for
pt$\to$ln.

\paragraph{Facet decomposition and the oracle account.} \texttt{store\_type} is the only
coarse facet our symbolic rule cannot beat its majority class on, so it isolates the part of
the coarse task that field derivability cannot explain. CORD \texttt{store\_type} accuracy:
\rzero{} $.869$, \rone{} $.828$, \rtwo{} $.869$, \rthree{} $.838$, \rthreeb{} $.848$,
\rfour{} $.899$, \rlndir{} $.879$, \rptdir{} $.849$, \rfourn{} $.889$, \rfoursh{} $.586$,
\rpipe{} $.929$. Paired against \rfour{}: \rzero{} $+3.0$ $[-3.0,+9.1]$, \rtwo{} $+3.0$
$[-4.0,+10.1]$, \rthree{} $+6.1$ $[-1.0,+13.1]$, \rfourn{} $+1.0$ $[-5.1,+7.1]$, \rone{}
$+7.1$ $[1.0,14.1]$, \rfoursh{} $+31.3$ $[21.2,41.4]$, \rpipe{} $-3.0$ $[-8.1,+1.0]$.
WildReceipt \texttt{store\_type}: \rzero{} $.840$, \rone{} $.830$, \rtwo{} $.870$,
\rthree{} $.860$, \rfour{} $.910$, \rfourn{} $.860$, \rptdir{} $.870$, \rlndir{} $.910$,
\rfoursh{} $.590$, \rpipe{} $.940$; paired against \rfour{}: \rthree{} $+5.0$ $[1.0,10.0]$,
\rfourn{} $+5.0$ $[1.0,10.0]$, \rone{} $+8.0$ $[1.0,15.0]$, \rzero{} $+7.0$ $[0.0,15.0]$,
\rtwo{} $+4.0$ $[0.0,9.0]$, \rpipe{} $-3.0$ $[-7.0,+1.0]$. On CORD the oracle account is not
excluded; on WildReceipt it is.

\paragraph{Neutral-content arm.} \rfourn{} is built by replacing only the conditioning
slot of every \rfour{} training example, so the templates are byte-identical outside that
slot. The hypothesis slot becomes \texttt{store\_type=unknown, payment\_method=unknown,
has\_discount=unknown, has\_surcharge=unknown}: \texttt{unknown} rather than a real value
such as \texttt{has\_discount=no}, which would be false for many receipts and would make
the arm adversarial rather than neutral. The field-list slot keeps its line count and its
\texttt{- category: text} shape with every category and text replaced by a constant
placeholder, since masking only the texts would leave the categories in place and the
presence of a \texttt{sub\_total.tax\_price} line already implies
\texttt{has\_surcharge=yes}. Training and evaluation follow the \rfour{} recipe unchanged.
CORD: point $.787$, line $.904$; \rfour{}$-$\rfourn{} point $+2.2$ $[0.6,4.0]$, line $+4.3$
$[1.5,6.8]$; \rfourn{}$-$\rthree{} point $+0.9$ $[-0.7,+2.5]$, line $+1.0$ $[-0.3,+2.5]$.
WildReceipt: point $.610$, line $.888$; \rfour{}$-$\rfourn{} point $-1.9$ $[-7.0,+2.8]$,
line $+0.3$ $[-2.8,+3.2]$; \rfourn{}$-$\rthree{} point $+8.3$ $[4.2,12.8]$, line $+1.0$
$[-0.5,+2.5]$.

\paragraph{Format-collapse control.} Because single-task tuning specializes the output
format, we separate malformed output from wrong output using the saved generations. On the
CORD line task  \rone{} parses $99/99$ with all four keys present and all values inside the
allowed set (zero-shot also $99/99$), so its $-5.3$ loss contains no formatting component.
On the CORD point task  \rtwo{} parses $93/99$ against zero-shot's $97/99$; over the $92$
receipts both parsed,  \rtwo{}$-$ \rzero{} is $-5.5$ $[-8.3,-2.9]$ against $-7.6$
$[-11.0,-4.4]$ on all $99$. On WildReceipt the same correction moves  \rtwo{}$-$ \rzero{}
from $-2.7$ $[-6.0,+0.3]$ to $-0.6$ $[-1.9,+0.8]$ over the $86$ receipts both parsed, so we
do not claim fine-side damage there.

\paragraph{Seed-aware intervals.} Each replicate resamples training seeds with
replacement independently per arm and test samples with replacement, so the interval carries
both sources of variance (single-seed $\to$ seed-aware). CORD: \rfour{}$-$\rtwo{} line
$+4.8$ $[2.3,7.3]\!\to\!+6.0$ $[2.5,9.4]$; \rfour{}$-$\rthree{} line $+5.3$
$[2.5,8.1]\!\to\!+5.7$ $[2.5,8.8]$; \rfour{}$-$\rzero{} line $+3.5$
$[1.0,6.1]\!\to\!+3.5$ $[1.0,6.0]$; \rfour{}$-$\rone{} point $+0.5$
$[-1.8,+2.6]\!\to\!+1.6$ $[-0.7,+4.0]$; \rfour{}$-$\rthree{} point $+3.0$
$[1.1,5.2]\!\to\!+2.2$ $[0.3,4.3]$. WildReceipt: \rfour{}$-$\rthree{} point $+6.4$
$[1.8,11.1]\!\to\!+5.1$ $[0.9,9.6]$; \rfour{}$-$\rone{} point $+13.8$
$[7.6,19.9]\!\to\!+14.9$ $[8.8,21.7]$; \rfour{}$-$\rthree{} line $+1.2$
$[-1.5,+4.0]\!\to\!+1.7$ $[-1.5,+4.9]$; \rfour{}$-$\rzero{} line $+5.5$
$[1.8,9.2]\!\to\!+4.7$ $[1.4,8.3]$; \rfour{}$-$\rtwo{} line $-1.2$
$[-4.5,+1.8]\!\to\!-0.8$ $[-3.9,+2.3]$. Every interval that excluded zero still excludes
it and every interval that spanned zero still spans it. \rtwo{} has two CORD seeds and
\rzero{} is deterministic, so those contrasts resample one arm only.

\paragraph{Probe control task and the skewed facet.} The control task permutes the
training labels, preserving each facet's marginal. That is the standard construction but it
is uninformative on a facet whose majority class dominates: for \texttt{has\_discount}
(majority $.899$) the permuted control reaches $.853$--$.873$ across regimes, so
$\mathrm{acc}-\mathrm{acc}_{\text{ctrl}}$ collapses to $\le .04$ by construction. With
labels drawn uniformly over the two classes instead, selectivity is $+.36$ to $+.42$. Held-out
accuracy, balanced accuracy and the permuted control for the facet:
\rzero{} $.909/.772/.873$, \rone{} $.899/.633/.869$, \rtwo{} $.909/.683/.873$, \rthree{}
$.879/.578/.853$, \rfour{} $.859/.566/.873$. The facet is decodable above chance in every
regime and \rfour{} is not above \rzero{} on it, consistent with the other three facets.

\paragraph{Probe cross-regime contrasts.} Paired over the 99 test receipts at the
cross-validated layer, \texttt{store\_type}: \rfour{}$-$\rzero{} $+2.0$ $[-4.0,+8.1]$,
\rfour{}$-$\rthree{} $+9.1$ $[1.0,17.2]$.

\paragraph{Mechanism-battery intervals.} Intervention sensitivity, direct paired
contrasts against zero-shot: \rone{} $+2.6$ $[-0.2,+5.3]$, \rthree{} $+2.6$
$[-0.5,+6.2]$, \rfour{} $+2.1$ $[-0.3,+4.6]$. Gold hypothesis against plain prompt,
within model: \rone{} $+1.0$ $[-0.7,+2.8]$, \rthree{} $+1.1$ $[-0.6,+3.2]$, \rfour{}
$-0.4$ $[-1.6,+0.8]$. Modality, fields-only minus image-only: \rfour{} $+1.0$
$[-1.5,+3.8]$, all other regimes negative; \rfour{} minus zero-shot difference of gaps
$+2.8$ $[-1.0,+6.3]$. Attribution, paired difference of median lift \rfour{} minus
zero-shot: \texttt{store\_type} $+0.59$ $[0.30,0.78]$, \texttt{has\_surcharge} $+0.45$
$[0.26,0.69]$. 4B contrasts: \rfour{}$-$\rthree{} point $-2.0$ $[-4.5,+0.4]$,
\rfour{}$-$\rone{} point $-0.5$ $[-3.1,+2.2]$, \rfour{}$-$\rzero{} line $+2.8$
$[+0.5,+5.1]$, \rfour{}$-$\rtwo{} line $+17.2$ $[12.9,21.5]$.

\paragraph{Probe robustness.} Probes are scikit-learn logistic regressions with L2 regularisation at $C=0.5$, a maximum of 2000 iterations, and standardised features, fitted on the last-token hidden state of every fourth decoder layer. Cross-validated (5-fold, train-side) layer selection with 20
train-bootstrap refits; accuracy at the held-out layer (refit sd): \texttt{store\_type}
\rzero{} $.748$ ($.022$), \rone{} $.707$ ($.037$), \rtwo{} $.737$ ($.026$), \rthree{}
$.677$ ($.029$), \rfour{} $.768$ ($.027$); \texttt{payment\_method}
$.929/.929/.879/.879/.889$ (sd $.020$--$.027$); \texttt{has\_surcharge}
$.929/.899/.939/.919/.939$ (sd $.014$--$.017$). Test-side best-layer selection inflates
individual cells by up to $.12$ (\rthree{} \texttt{store\_type}), so the main text
reports the held-out numbers.

\paragraph{Symbolic rule baseline.} An oracle over gold point fields, using the
annotation guide's constraint rules (CORD: cash/change fields $\to$ cash, credit card
field $\to$ card, e-money field $\to$ e\_money, several $\to$ mixed, none $\to$
unknown; discount or void fields $\to$ discount; tax or service fields $\to$ surcharge;
majority class for \texttt{store\_type}) reaches per-facet accuracy
$.424/.929/.919/.970$ and line accuracy $.811$ on CORD test. On WildReceipt only
\texttt{has\_surcharge} is field-derivable (tax or tips present $\to$ surcharge,
$.860$; all other facets at majority class), for line accuracy $.690$. Both sit below
zero-shot ($.912$/$.835$), and \rfour{} exceeds the oracle by $13.6$ and $20.0$ points.

\paragraph{Self-conditioning at test time.} Conditioning \rfour{} on its own predicted
fields lifts CORD line accuracy from $.947$ to $.955$ (gold fields $.965$); on the
WildReceipt point task its own predicted hypotheses give $.550$ and gold hypotheses
$.551$, both below the plain prompt's $.591$.

\section{Pipeline Baseline}
\label{app:pipe}

The two-stage pipeline replaces \rfour{}'s gold conditioning context with a stage-1 model's
predicted context, at both training and test. Stage-1 is the trained single-task model
(\rone{} for fields, \rtwo{} for facets); stage-2 is a fresh LoRA trained, with prompts
byte-identical to \rfour{}'s, on those predictions rather than gold labels. At test the same
stage-1 model produces the context for stage-2. Stage-1 predictions on the training set match
test-set quality (CORD line $.888$, point $.813$; WildReceipt line $.905$, point $.408$), so
the pipeline conditions on genuinely predicted, not memorized, context; on WildReceipt the
point extractor degenerates on roughly a third of receipts, and stage-2 learns to fall back
on the image, which is why the pipeline's line side is unharmed while its point side
collapses. Per-facet WildReceipt line accuracy for the pipeline is
$.94/.80/.95/.93$ (\texttt{store\_type}/\texttt{payment}/\texttt{discount}/\texttt{surcharge}).

\section{Attribution and Modality}
\label{app:attrib}

\begin{figure*}[t]
\centering
\includegraphics[width=0.98\textwidth]{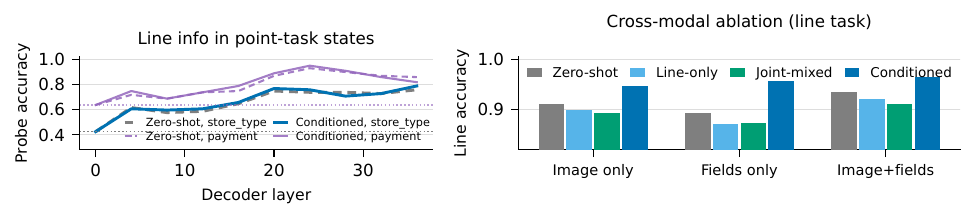}
\caption{Left: layer-wise probe accuracy for line information in point-task states, zero-shot
(dashed) vs \rfour{} (solid), dotted lines mark the majority baselines ($.42$ \texttt{store\_type}, $.64$
\texttt{payment\_method}); decodability rises with depth. Right: line accuracy by input modality. \rfour{} is the only model
for which the field text alone beats the image alone, and fusion is superadditive for all.}
\label{fig:mech}
\end{figure*}

The remaining two instruments both run the line task \emph{with fields in context}, which
is \rfour{}'s own training template and out of distribution for every other regime. We
report them with that asymmetry stated, because it is a live alternative explanation for any
gap they show. We backpropagate each facet answer's log-probability to the token embeddings
and aggregate the gradient norm over each field's span; writing $a(e)$ for the mean
attribution density on field $e$, the evidence lift for facet $f$ is
\begin{equation}
\label{eq:lift}
\mathrm{lift}_f=\frac{\tfrac{1}{|E_f|}\sum_{e\in E_f} a(e)}
{\tfrac{1}{|Y_{\pt}|}\sum_{e\in Y_{\pt}} a(e)},
\end{equation}
the density on annotated evidence relative to all fields, with $\mathrm{lift}=1$ as the
count-matched null. Two checks discipline it (Table~\ref{tab:attribval}): a float32
recomputation moves the surviving facets by at most $.07$ while shifting the twelve-sample
\texttt{has\_discount} by $.36$, and a lexical-overlap null recomputes lift on evidence
sharing no tokens with the answer, which \texttt{payment\_method} fails (68\% overlap,
residual $1.00$) and \texttt{store\_type} and \texttt{has\_surcharge} survive.

\begin{figure}[t]
\centering
\includegraphics[width=0.98\columnwidth]{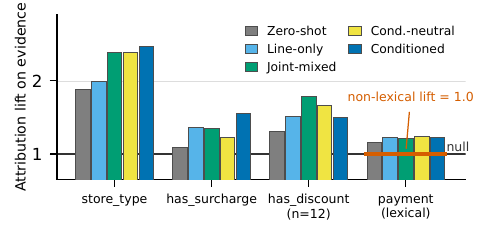}
\caption{Median attribution lift on annotated evidence fields (1.0 = count-matched null).
\texttt{payment\_method} is shown for completeness but fails the lexical-overlap null;
\texttt{has\_discount} has $n{=}12$. Lift is above the null for every regime; \rfour{} is highest on \texttt{has\_surcharge}, level with \rthree{} on \texttt{store\_type} and below it on \texttt{has\_discount}. \rfourn{} bounds the format contribution: it reproduces most of \rfour{}'s \texttt{store\_type} gain over zero-shot and little of its \texttt{has\_surcharge} gain.}
\label{fig:attrib}
\end{figure}

The neutral-content control bounds how much of this is format exposure: its median lift is $2.385$ on \texttt{store\_type} and $1.231$ on \texttt{has\_surcharge}, so it reproduces $86\%$ of \rfour{}'s gain over zero-shot on the first facet and $29\%$ on the second, and \rfour{} exceeds it on both by paired median differences of $+0.139$ $[0.040,0.293]$ and $+0.222$ $[0.060,0.420]$. On the surviving facets every model attributes above the null and \rfour{} attributes most sharply (\texttt{store\_type} $1.88\!\to\!2.47$, \texttt{has\_surcharge}
$1.10\!\to\!1.55$; paired median-lift differences over zero-shot $+0.59$ $[0.30,0.78]$ and
$+0.45$ $[0.26,0.69]$), and its top attributed field is annotated evidence for $.87$ of
samples against a $.17$ chance rate. The instrument does not, however, separate \rfour{}
from \rthree{}, which is level with it on \texttt{store\_type} and above it on
\texttt{has\_discount} (Figure~\ref{fig:attrib}), so attribution sharpness cannot by
itself explain why \rthree{} fails behaviourally. Because the links come from LLM judges we
read the alignment as convergent validity rather than as validation of either side.

Under three input conditions, image only, gold field text only, and both
(Figure~\ref{fig:mech}, right), fusion is superadditive for every model, and the regimes
separate on the text channel: field text alone is the weaker channel for zero-shot,
\rtwo{} and \rthree{} ($.87$--$.89$, below image-only) and the stronger one for \rfour{}
($.957$ vs $.947$, $+1.0$ paired, $[-1.5,+3.8]$). The direction is consistent across facets
but the gap is not resolvable, and format familiarity predicts the same sign, so we report
the channel reversal as a description of \rfour{}'s behaviour rather than as evidence about
its representations.

\section{Third Domain: FUNSD}
\label{app:funsd}

FUNSD is 199 scanned 1990s business documents (fax cover sheets, memoranda, progress reports,
questionnaires) in the official 149/50 split. The fine task reuses the corpus's own gold entity
annotations, decoded from word-level BIO into \texttt{(category, text)} pairs over
\{header, question, answer\} and scored with the same multiset F1; a median document carries 36
entities. The receipt facets do not transfer, so the coarse task uses four new ones with the same
shape as the receipt schema, one semantic facet judged holistically and three structural facets
anchored in specific field types: \texttt{doc\_type} (fax\_cover, memo\_letter, report,
form\_questionnaire, other), \texttt{has\_checkbox}, \texttt{has\_cc\_field} and
\texttt{has\_unfilled\_field}. Gates, success criteria and a declared direction prediction were fixed before any label was aggregated. Both pre-registered outcomes are reported as they fell. The primary criterion, conditioned line accuracy at least equal to every other trained regime, passes on the pre-registered regime set ($.920$ against $.770$, $.905$ and $.810$) and fails once the post-hoc \rlndir{} ablation counts ($.935$), exactly as on WildReceipt. The declared direction prediction is refuted: we predicted the gain on the side with more headroom, here the fine side at a zero-shot F1 of $.117$ against a coarse accuracy of $.915$, and it appeared on the coarse side. We report it refuted rather than reinterpreting the ordering afterwards. The $50$ test documents are human-validated in full at $91.0\%$ agreement.

\paragraph{Seeds and recipe sweep.} Table~\ref{tab:fssweep} reports every FUNSD arm we trained. Three seeds separate the two central arms on the coarse side without overlap, and the paired contrasts at seeds 1 and 2 are $+16.0$ $[11.0,21.0]$ and $+15.0$ $[10.0,20.5]$ on line, $+44.0$ $[30.0,58.0]$ and $+46.0$ $[32.0,60.0]$ on \texttt{doc\_type}. The fine side does not separate consistently: $-1.1$ $[-8.1,+5.9]$ at seed 1 and $+7.3$ $[0.8,14.1]$ at seed 2. Pooling the seeds into the two-level bootstrap used for the other two corpora, which resamples seeds and then documents, gives \rfour{}$-$\rthree{} line $+15.3$ $[10.7,20.3]$ and \texttt{doc\_type} $+46.0$ $[32.0,60.0]$, and \rfour{}$-$\rfourn{} line $+18.9$ $[13.2,25.2]$ with the same \texttt{doc\_type} interval, so the coarse-side result does not rest on the single run in Table~\ref{tab:funsd}. The recipe sweep is reported in Section~\ref{sec:results}; the two arms that move it are mixed training at $2{\times}10^{-5}$, which recovers, and at four epochs, which degrades further, so the direction of the recipe effect is not simply an amount of training.

\paragraph{Mixing ratio.} Repeating the line rows of the mixed training set two and four times, changing nothing else, lifts macro coarse accuracy from $.770$ to $.805$ in both arms ($+3.5$, $[1.0,6.5]$) through the three structural facets, and leaves \texttt{doc\_type} at the majority label for all $50$ documents in both, $+0.0$ $[0.0,0.0]$ against the balanced arm. The fine side pays for the reweighting, $.426$ falling to $.359$ and $.369$. Both arms stay $11.0$ coarse points and $38.0$ \texttt{doc\_type} points below the untuned model, so the collapse is not an artefact of the task proportion.

\paragraph{The predicate at the second learning rate.} Read against single-task models retrained at $2{\times}10^{-5}$ rather than at $10^{-4}$, mixed training still does not reinforce: it is above matched \rone{} on the fine side ($+3.1$, $[-1.4,+7.9]$) and below matched \rtwo{} on the coarse side ($-2.5$, $[-5.5,+0.5]$), so it trades there as it does at the shared rate. Conditioning does not reinforce at that rate either, sitting below both matched arms ($-6.0$ $[-13.8,+1.5]$ fine, $-0.5$ $[-3.5,+2.5]$ coarse), and mixed training leads it there on the fine side by $+9.1$ $[2.2,16.3]$, the only arm in the study that resolvably beats conditioning on either side of any corpus. The four arms cluster on the coarse side, $.890$ to $.926$ against zero-shot's $.915$, which is why both the collapse and its repair are reported as properties of the shared recipe on this corpus.

\begin{table*}[t]
\centering\small\setlength{\tabcolsep}{9pt}
\begin{tabular}{lccc}
\toprule
FUNSD arm & Point F1 & Line Acc & \texttt{doc\_type} \\
\midrule
\rthree{} seeds 0/1/2 & .426/.401/.372 & .770/.755/.765 & .520/.520/.520 \\
\rthree{} lr $2{\times}10^{-5}$ & .417 & .890 & .880 \\
\rthree{} four epochs & .433 & .700 & .520 \\
\rthree{} line rows $2\times$ & .359 & .805 & .520 \\
\rthree{} line rows $4\times$ & .369 & .805 & .520 \\
\midrule
\rfour{} seeds 0/1/2 & .430/.390/.445 & .920/.915/.915 & 1.000/.960/.980 \\
\rfour{} lr $2{\times}10^{-5}$ & .325 & .910 & .960 \\
\midrule
\rfourn{} seeds 0/1 & .402/.369 & .750/.705 & .520/.520 \\
\midrule
\rone{} lr $2{\times}10^{-5}$ & .386 & .920 & .920 \\
\rtwo{} lr $2{\times}10^{-5}$ & .165 & .915 & .960 \\
\bottomrule
\end{tabular}
\caption{FUNSD seed replications and recipe sweep, all at $n{=}50$. Seed 0 is the run reported in Table~\ref{tab:funsd}. The last block gives the single-task arms retrained at the second rate, without which the arms at that rate have no matched reference. Nothing is bolded here: the arms differ in recipe as well as in regime, so a single best-in-column would compare across two changes at once.}
\label{tab:fssweep}
\end{table*}

The committee gate passed on both splits: per-facet majority $100\%$ everywhere, Fleiss $\kappa$
$.734$--$.957$ on the test split and $.754$--$.910$ on the training split, and no facet with a
majority class above $90\%$. Three annotators then relabelled all 50 test documents blind, at
$91.0\%$ agreement with the committee ($\kappa$ $.96$ \texttt{has\_checkbox}, $.89$
\texttt{has\_unfilled\_field}, $.79$ \texttt{doc\_type}, $.57$ \texttt{has\_cc\_field});
\texttt{has\_cc\_field} is the least reliable facet and its low $\kappa$ reflects both genuine
ambiguity and an $82\%$ base rate. Training reuses the CORD recipe unchanged. One caveat is
specific to this corpus: because the facet space is coarse, a shuffled hypothesis coincides with
the document's own labels for $8.7\%$ of training examples, so the shuffled arm is a weaker
control here than on the receipts; the field-list slot never coincides.

\section{Second Backbone (Qwen3-VL-4B)}
\label{app:4b}

\begin{figure}[h]
\centering
\includegraphics[width=0.98\columnwidth]{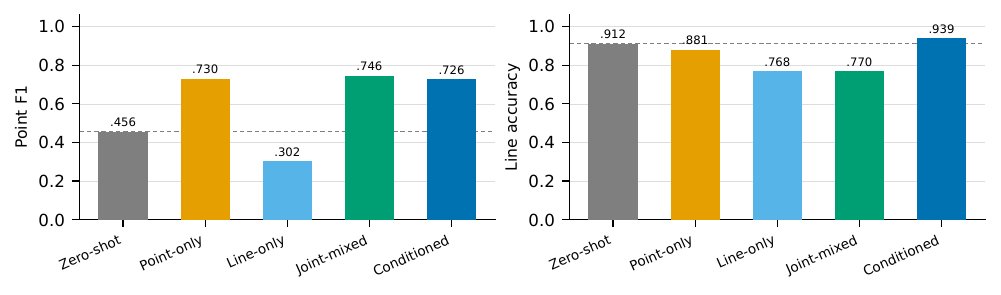}
\caption{Five-regime replication on Qwen3-VL-4B, CORD test, identical recipe. Line-side
pattern amplifies: \rtwo{} and \rthree{} fall 14 points below zero-shot on the line task
itself, \rfour{} alone stays above it. Point side: \rthree{} $.746$ exceeds \rfour{} $.726$ ($-2.0$ $[-4.5,+0.4]$) and
\rone{} $.730$ sits $0.5$ above it ($[-3.1,+2.2]$), so the point-side advantage of
conditioning is specific to 8B and its point-side no-loss holds only within noise at 4B;
the line advantage holds at both scales ($+2.8$ $[0.5,5.1]$ over zero-shot).}
\label{fig:4b}
\end{figure}

The 4B recipe reuses 8B hyperparameters without tuning; the amplified self-damage of
\rtwo{}/\rthree{} at 4B may partly reflect that choice, which does not affect the ordering
conclusion.

\section{Zero-Shot Pilot and Iterative Loop}
\label{app:pilot}

Before training we measured the same task pair zero-shot on 198 CORD training receipts.
Single-pass joint prompting interferes (line $.869$ vs $.897$ single-task; $-7.7$ points on
\texttt{store\_type}); an iterative loop, hypothesis $\to$ conditioned extraction $\to$
field-conditioned re-judgment, recovers part of it (line $+2.15$ over joint, $[0.1,4.3]$),
with the gain coming from the hypothesis-to-extraction direction ($+2.4$ point F1) while the
reverse direction was flat and a second iteration degraded. Training-time results in the main
text mirror this inference-time asymmetry. The loop costs four generations against one for
joint prompting; we did not run a compute-matched sampling control, and treat the pilot as
motivation rather than a method claim.

\section{Activation Patching (Inconclusive)}
\label{app:patch}

We attempted a representation-level test: transplant the last-prompt-token residual state at
layers 16/20/24/28 (and the band) from the \rfour{} model into \rtwo{} or zero-shot during a
scored five-way \texttt{store\_type} choice, with a random-donor control. No configuration
moved recipient accuracy or agreement with the donor. The scoring protocol itself proved
unreliable, the donor's forced-prefix accuracy ($.495$) falls far below its free-generation
accuracy ($.899$), so this null is uninformative about localization in either direction, and
no representation-level conclusion is drawn from it.

\section{Annotation Pipeline and Prompts}
\label{app:prompts}

\paragraph{Committee.} Each judge receives the receipt image and the gold field list
(id, category, text per instance), and returns the four facets, per-facet evidence field
ids, a constraint-applicability flag, and per-facet confidences as JSON. Judges never see
one another's outputs; aggregation is per-facet majority vote, and evidence links keep the
ids named by at least two judges. Per-sample cost averaged \$0.014 across the three judges.

\paragraph{Judge system prompt (verbatim).}
\begin{footnotesize}
\begin{verbatim}
You are annotating a receipt image for a
document-understanding dataset. You are
given the receipt image and its gold
extracted fields (each with id, category,
text). Return ONLY a JSON object with
exactly these keys:
{"store_type":
   "restaurant|cafe_beverage|
    bakery_dessert|retail_convenience
    |other",
 "payment_method":
   "cash|card|e_money|mixed|unknown",
 "has_discount": "yes|no",
 "has_surcharge": "yes|no",
 "evidence": {"store_type": [],
   "payment_method": [],
   "has_discount": [],
   "has_surcharge": []},
 "constraint_table_applies": true,
 "confidence": {...}}
Definitions:
- store_type: judge HOLISTICALLY from
  store name, menu-item semantics, and
  overall layout. cafe_beverage =
  primarily drinks;
  bakery_dessert = baked goods / desserts;
  retail_convenience = packaged goods /
  non-prepared food retail; restaurant =
  prepared meals; other = unclear or none.
- payment_method: cash (cash/change
  fields), card (credit/debit), e_money,
  mixed
  (multiple methods), unknown.
- has_discount: yes if any discount, void,
  promo, or negative amount reduces
  the bill.
- has_surcharge: yes if tax and/or service
  charge is ADDED to the bill.
- evidence: for each facet, the field ids
  (ONLY from the provided list) supporting
  your value. Use [] when no listed field
  supports it (e.g. store_type judged from
  the image alone).
- constraint_table_applies: true iff these
  expectations hold on this receipt:
  payment=cash implies cash & change
  fields present; has_discount=yes implies a
  discount/void field present;
  has_surcharge=yes implies a tax/service
  field present.
Rules: NEVER invent field ids. Judge only
from the image and the provided fields.
Output raw JSON, no markdown.
\end{verbatim}
\end{footnotesize}

\paragraph{Task prompts (verbatim).} Point:
\begin{footnotesize}
\begin{verbatim}
Extract ALL key information fields from
this receipt image.
Allowed categories: {ontology}.
Return ONLY a JSON array, one entry per
field instance in reading order:
[{"category": "...", "text": "..."}].
Copy text EXACTLY as printed.
\end{verbatim}
\end{footnotesize}
Line:
\begin{footnotesize}
\begin{verbatim}
Classify this receipt on 4 document-level
facets. Return ONLY JSON:
{"store_type": "...", "payment_method":
 "...", "has_discount": "yes|no",
 "has_surcharge": "yes|no"}
store_type: judge holistically (store
name, item semantics, layout).
payment_method:
how the bill was paid. has_discount: any
discount/void/promo reduces the bill.
has_surcharge: tax and/or service charge
added.
\end{verbatim}
\end{footnotesize}

\paragraph{Conditioned-training templates ($g$, $h$ in Eq.~\ref{eq:r4}, verbatim).}
Point side prefix:
\begin{footnotesize}
\begin{verbatim}
Document-level hypothesis for this
receipt: store_type=...,
payment_method=...,
has_discount=..., has_surcharge=... .
Use it as context (e.g. expected payment/
discount/surcharge fields), but trust the
image over the hypothesis.
\end{verbatim}
\end{footnotesize}
Line side prefix:
\begin{footnotesize}
\begin{verbatim}
Fields extracted from this receipt:
- category: text
- ...
Using BOTH the image and these fields,
classify this receipt on 4 document-level
facets. ...
\end{verbatim}
\end{footnotesize}
\rfoursh{} uses byte-identical templates whose hypothesis and field list come from document
$\pi(i)$, a deterministic shift of the training index.

\section{Dataset Examples}
\label{app:examples}

\begin{figure}[h]
\centering
\includegraphics[height=1.85in]{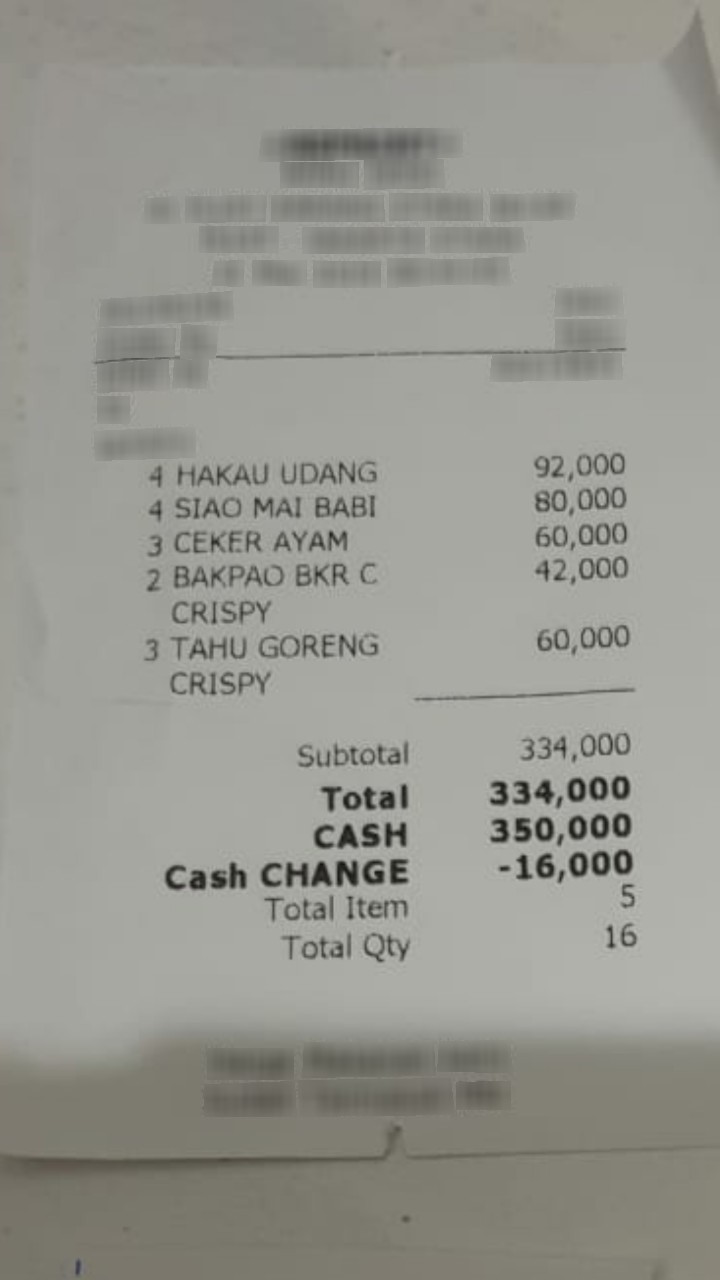}\hspace{2mm}
\includegraphics[height=1.85in]{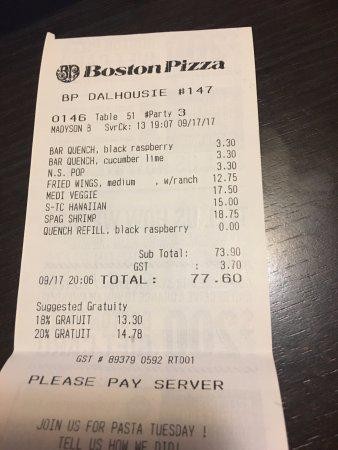}
\caption{Example receipts: CORD (left, Indonesian) and WildReceipt (right, English).}
\label{fig:examples}
\end{figure}

\noindent Doc-MRE record for the CORD example (fields abridged):
\begin{footnotesize}
\begin{verbatim}
fields: [
 {"id":"menu.nm#0","text":"1PC KFI"},
 {"id":"menu.cnt#0","text":"1"},
 {"id":"menu.price#0","text":"12,500"},
 ... 8 more field instances ...
 {"id":"total.total_price#0",
  "text":"25,000"},
 {"id":"total.cashprice#0",
  "text":"50,000"},
 {"id":"total.changeprice#0",
  "text":"25,000"}]
line labels: {"store_type":"restaurant",
 "payment_method":"cash",
 "has_discount":"no","has_surcharge":"no"}
evidence: {"store_type":["menu.nm#0"],
 "payment_method":["total.cashprice#0",
   "total.changeprice#0"],
 "has_discount":[], "has_surcharge":[]}
\end{verbatim}
\end{footnotesize}
WildReceipt records use the twelve-category schema (\texttt{store\_name},
\texttt{prod\_item}, \texttt{subtotal}, \texttt{tax}, \texttt{tips}, \texttt{total}, etc.);
document-level distributions differ markedly from CORD (payment unknown on half the
receipts, surcharge present on 68\%), which is what makes it a useful second schema.

\section{Human Annotation: Guide and Interface}
\label{app:guide}

\begin{figure}[h]
\centering
\includegraphics[width=0.98\columnwidth]{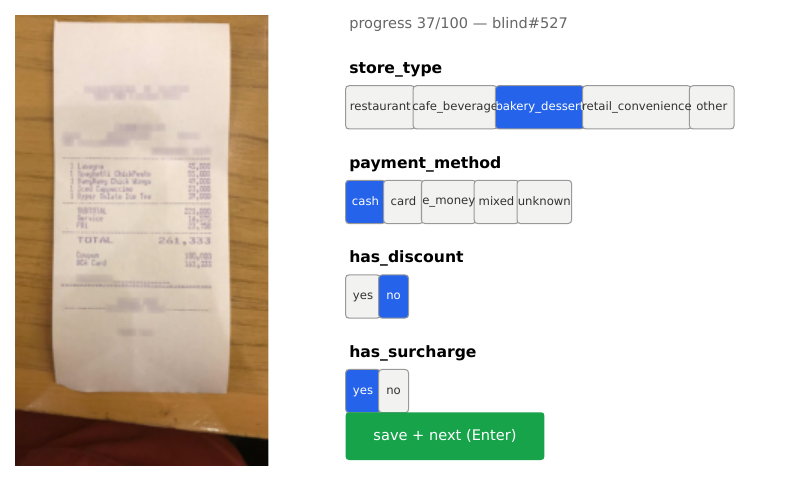}
\caption{The annotation interface (blind mode): receipt image with click-to-zoom on the
left; the four facet button groups, keyboard shortcuts, and save-and-next on the right. In
blind mode no committee label is ever displayed and options start unselected.}
\label{fig:ui}
\end{figure}

The guide given to annotators, translated from the working copy:

\paragraph{Protocol.} Judge each receipt independently from the image alone; accuracy over
speed (20--40 seconds per receipt); all four facets required before saving; no discussion
of individual samples.

\paragraph{Decision rules.}
\texttt{store\_type}: judge from store name, items sold, and layout together; mixed
businesses by the dominant category on the ticket (a coffee shop selling one cake is still
\texttt{cafe\_beverage}); when name and items conflict, items win.
\texttt{payment\_method}: only lines showing the actual payment count; advertised payment
options do not; a change line implies cash; QRIS counts as e-money.
\texttt{has\_discount}: any discount, promo, voucher, void, or negative amount; a
zero-amount discount line counts as no.
\texttt{has\_surcharge}: any separately listed tax (TAX/PAJAK/PB1/PPN) or service charge;
``prices include tax'' with no separate line counts as no.

\paragraph{Vocabulary.} Indonesian receipt terms: TUNAI = cash, KEMBALI/KEMBALIAN = change,
PB1/PPN = tax, DISKON/POTONGAN = discount, and OVO/GoPay/DANA/ShopeePay/QRIS are e-wallets.

\paragraph{Fallbacks.} If undecidable: \texttt{store\_type} = other,
\texttt{payment\_method} = unknown; for the two binary facets, absence of visible evidence
means no.

\section{Pre-registration of the WildReceipt Replication}
\label{app:prereg}

Before any WildReceipt annotation was aggregated we fixed: a quality gate (three-judge
majority $\ge 80\%$ and $\kappa\ge.70$ on every facet, else the dataset is excluded with the
gate reported), the success criteria (primary: \rfour{} line $\ge$ every other trained
regime, direction consistent with CORD; secondary: \rfour{} point $\ge$ \rthree{}), and the
analysis plan (CORD recipe unchanged, seed 0, paired bootstrap CIs). The gate passed
(majority 100\% on all facets, $\kappa$ .830--.948). The secondary criterion passed
(\rfour{} point $+6.4$ over \rthree{}, $[1.8,11.1]$). The primary criterion passed on the
pre-registered regime set (\rfour{} line $.890$ against \rthree{} $.878$ and \rone{}
$.785$), but \emph{fails} once the post-hoc arms are counted: \rtwo{} reaches $.903$ and
\rpipe{} $.905$, both above \rfour{}. We therefore report the primary criterion as failed
and restate the WildReceipt finding as follows: \rfour{} matches the best line regime
within the paired interval ($-1.5$ against \rpipe{}, $[-4.5,+1.5]$) and leads every
pre-registered single-forward-pass regime on point, while two post-hoc arms, \rfourn{}
($.610$) and \rptdir{} ($.596$), are nominally higher on point with unresolvable differences
($-1.9$ $[-7.0,+2.8]$ and $-0.5$ $[-5.0,+3.8]$). The direction-flip observation is labeled post-hoc.

\end{document}